\documentclass[11pt]{article}

\usepackage[T1]{fontenc}
\usepackage[utf8]{inputenc}
\usepackage[english]{babel}

\usepackage[top=2.5cm,bottom=2.5cm,left=3cm,right=3cm]{geometry}

\usepackage[protrusion=true,expansion=false]{microtype}

\usepackage{xcolor}

\usepackage{amsmath,amssymb}

\usepackage{graphicx}
\graphicspath{{./}{figures/}}
\usepackage[section]{placeins}     
\usepackage{booktabs}
\usepackage{threeparttable}

\usepackage{siunitx}
\usepackage{natbib}
\usepackage{url}                    
\usepackage[hidelinks]{hyperref}     

\hypersetup{
  pdftitle  = {Detecting Authorship in Political Texts with Inductive Stylometry},
  pdfauthor = {Gennadii Iakovlev and Levente Littvay},
  pdfkeywords = {stylometry, political communication, authorship attribution, text as data, UMAP},
  pdfproducer = {LaTeX with hyperref},
  pdfcreator  = {pdflatex}
}

\title{Detecting Authorship in Political Texts with Inductive Stylometry}
\author{Gennadii Iakovlev\thanks{ORCID: 0000-0002-8355-6914} \and Levente Littvay\thanks{ORCID: 0000-0003-2022-6886}}
\date{September 2026}

\begin{document}
\maketitle

\begin{abstract} \noindent Political texts are rarely authored by the nominal
	speaker alone. Tweets, speeches, reports, and official statements are
	drafted, edited, or harmonized by staff, yet political science has paid
	limited attention to the stylistic traces these hidden authors leave
	behind. This paper develops and stress-tests an inductive stylometric
	approach for recovering latent authorship structure in political
	communication, combining character 3-gram features
	with UMAP dimensionality reduction, and Burrows' Delta. We apply the approach to six
	corpora that vary in length (from tweets to long documents), in mode
	(written and oral), and in language (English and Hungarian).
	The approach recovers near-disjoint analyst fingerprints in formal
	legal prose in both languages, sorts a politician's tweets into
	validated subsets while uncovering additional insights, and
	distinguishes scripted from improvised speech. It fails, however, to
	resolve individual speechwriters within scripted corpora.
	Frequency-based stylometry is thus a powerful tool that, depending
	on authorial signal strength and institutional editing, can uncover
	authorship traces relevant to legislative studies, political
	communication, and policy research.
\end{abstract} \vspace{1em}

\setlength{\parindent}{0pt}
\setlength{\parskip}{6pt}

\section{Introduction}

Stylometry is a quantitative text analytical technique historically reserved for verifying authorship in disputed literary cases \citep{Stamatatos2009}. Perhaps the most famous modern application was the forensic identification of J.K. Rowling as the true author of the crime novel \textit{The Cuckoo's Calling}, unmasking her pseudonym ``Robert Galbraith'' through distinct syntactic fingerprints \citep{Juola2015}. Yet, despite the origins of quantitative stylometry traced to the identification of Hamilton as the author of twelve disputed Federalist papers \citep{Mosteller1964}, and the explosion of computational text analysis in political science since the turn of the millennium \citep{Grimmer2013, Wilkerson2017}, stylometric approaches have received little to no attention in the field.\footnote{For a rare exception, see \citealp*{Airoldi2007}} We argue that neglecting stylometry is a critical oversight, particularly given the increasing opacity surrounding the true authorship of political texts. Political analysis should therefore account for the staff and other contributors who draft speeches, court decisions, laws, and diplomatic documents.

Our goal is to apply and stress-test classical stylometric approaches in the
noisy domain of political communication, stretching their applicability across
three distinct directions: from written literary texts to delivered speeches,
from long-form documents to short social-media posts, and from English to the
morphologically complex non-Indo-European language of Hungarian.

Our first two case studies apply the same algorithm to Congressional Research Service (CRS) reports, focusing on the American Law topic subset, where a small set of named staff analysts author documents that share topic, audience, and a highly formal legal register---a high-homogeneity English-language benchmark. We exploit this corpus in two nested forms. We begin with the single most controlled slice, the short-report series, in which document format and length are held constant alongside topic and register, isolating analyst style as cleanly as the data allow; we then relax the format control and pool the full American Law subset to show that the same analyst signal survives across document formats.

Third, we examine a corpus of Hungarian Ombudsman reports to test whether stylometric methods can recover different authors among official administrative legal texts---a genre written in a highly formal register that typically circulates without named individual authors. These reports, unusually, record the staff rapporteur who drafted each document, allowing us to validate what the method recovers against known authors. We chose Hungarian as it comes from a different language family than most European languages, allowing us to assess whether the approaches tested generalize broadly. We also use a Hungarian language case study, below, testing speech authorship.

Fourth, we aimed to benchmark these methods---most often scrutinized on novella-sized or longer works---on the short-format texts that increasingly define modern political communication. Using Donald Trump's tweets from the Trump Twitter Archive \citep{trumptwitterarchive}, from his campaign launch through Election Day, we leveraged a known hardware distinction: tweets sent from an Android device (hypothesized to be Trump himself) versus an iPhone (hypothesized to be staffers) as a ground-truth proxy for authorship. This served to validate whether the stylometric signal persists in texts limited by character counts.

Our fifth case study sought to determine if stylometric approaches designed for static and written literary texts are applicable to speeches—texts that are convoluted by the "noise" of oral delivery and the editing process inherent to the collaborative nature of speechwriting. We utilized Donald Trump's 2016 campaign speeches, distinguishing between those delivered on and off the teleprompter. This allowed us to test if stylometry can discriminate between the "Institutional Voice" (scripted) and the "Individual Voice" (ad-libbed), and whether inductive clustering could reveal distinct authorial hands within the scripted corpus.

Finally, to examine cross-linguistic generalizability and test the method's inductive capacity in the absence of a validated ground truth, we applied the approach to a linguistically divergent case: the political speeches of Hungarian Prime Minister Viktor Orbán. 

Our findings suggest that while political texts are indeed "messy", stylometry offers significant analytical leverage. We find that the most basic stylistic clusters can effectively be established in both short and long political texts.
While clear, individual "speechwriter clusters" did not inductively emerge from the speeches (suggesting a high degree of homogenization in formal addresses), stylometry significantly sharpened the separation between ad-lib delivery and teleprompter speeches and other deductively hypothesized groups (e.g., Android vs. iPhone).

\section{Methods}

Our empirical strategy relies on classical frequency-based stylometry, the family of techniques developed in authorship-attribution research over the past half century \citep{Mosteller1964, grieve2007, koppel2009, Stamatatos2009}. We construct feature sets familiar to computational social scientists: simple text-level indicators (such as length and sentiment), character 3-grams, and Burrows' Delta on frequent words. We represent each document as a high-dimensional vector, reduce these representations to two dimensions with Uniform Manifold Approximation and Projection (UMAP) for visualisation, and assess whether the resulting layout aligns with known or hypothesised authorship signals. If the technique proves useful in cases where authorship is verifiable, we can be confident that it will also be useful as an inductive technique to detect stylometric patterns where authorship is unavailable.

\subsection{Dimensionality reduction techniques}

Frequency-based stylometry begins by converting texts into numerical form. Each document is represented as a vector of character trigram frequencies---sequences of three consecutive characters---yielding a document-by-feature matrix with several hundred columns in our corpora. This representation captures stylistic habits that are difficult to observe directly: spelling preferences, punctuation patterns, word endings, affixes, spacing, and other sub-word regularities. Documents with similar trigram patterns lie close together in this high-dimensional space. Such spaces can be analysed computationally, but visual inspection requires mapping the hundreds of original dimensions into two.

This is the purpose of dimensionality reduction. Unlike Principal Component Analysis, which extracts the strongest linear axes of variation, the methods most used for exploratory text visualization today are nonlinear and neighborhood-preserving: texts that are stylistically similar in the original space should remain close to one another in the plot.

The two most widely used methods of this kind are t-distributed stochastic neighbor embedding, or t-SNE \citep{vanderMaatenHinton2008}, and Uniform Manifold Approximation and Projection, or UMAP \citep{McInnesHealyMelville2018}.\footnote{For an overview of machine-learning dimensionality reduction techniques, see \citealp{marx2024}.} Both are unsupervised: they arrange documents purely from the numerical patterns in the feature matrix, without access to authorship labels---exactly the property inductive stylometry requires.

The mathematical foundations of the two methods differ: t-SNE models pairwise similarities between points as probabilities and preserves them in a lower-dimensional map, while UMAP assumes the data lie on or near a lower-dimensional manifold and preserves its internal geometry. In our corpora, UMAP produced visibly crisper authorship separation than t-SNE.

The key practical issue in dimensionality reduction is the balance between local structure (the immediate neighborhood of each document) and global structure (the broader arrangement of clusters across the corpus). Too much emphasis on local detail produces a fragmented, carpet-like map in which broad communities are difficult to distinguish; too much emphasis on global separation produces very clear clusters at the cost of compressing their internal structure.

In UMAP, this balance is controlled mainly by the number of neighbors: a smaller neighbor count weights local structure, a larger one weights global structure.\footnote{For our samples of 500--1000 cases, a local setup would usually mean 5--20 neighbors, while a more global setup would mean 50--200 neighbors. The appropriate value, however, varies from model to model.} For inductive stylometry we suggest emphasizing global communities with a relatively larger neighbor count, which makes the plots better suited for detecting broad authorship patterns; more local settings can serve as supplementary diagnostics when the internal structure of a cluster is of special interest.

\subsection{Classic stylometric analysis}

Our classic stylometric analysis relies on two complementary frequency-based approaches. The first is a character 3-gram model projected with UMAP, which operationalises the high-dimensional representation and dimensionality-reduction logic described above. The second is Burrows' Delta, a classic authorship-attribution measure based on frequent-word profiles. These two approaches form the main body of the analysis. Additional interpretable diagnostics---including sentiment, document length, sentence length, and related dictionary-based indicators---are reported in Appendix~D (Figures~A3--A10).

For the character 3-gram analysis, we construct document-by-feature matrices of trigram frequencies, trimming features that occur fewer than five times across the corpus, and project them into two dimensions using UMAP with a cosine distance metric \citep{McInnesHealyMelville2018}. Every corpus is projected directly from its trimmed trigram matrix, with no intermediate reduction. Following the local--global trade-off discussed above, we scale the number of nearest neighbours with corpus size, using larger values on the bigger corpora to emphasise global community structure over fine-grained local detail.\footnote{Nearest neighbours and minimum distance by corpus: CRS American Law (both the short-report and full subsets), 60 and 0.30; Hungarian ombudsman, 50 and 0.05; Trump tweets, 50 and 0.20; Orb\'an speeches, 15 and 0.05; the 23 Trump speeches, 5 and 0.05. On the two CRS corpora, whose reports share substantial templated boilerplate, we additionally apply an iterative outlier filter before measuring cluster quality---removing documents that fall beyond eight median absolute deviations on either UMAP axis or whose fifth-nearest-neighbour distance exceeds six times the median---as a guard against near-duplicate boilerplate; on the analysed American Law subsets it flagged no documents.}

Alongside the trigram--UMAP analysis, we compute Burrows' Delta on the most frequent words. Burrows' Delta is a standard stylometric distance measure that compares texts or groups of texts according to their relative use of frequent words, especially common function words such as articles, prepositions, pronouns, and conjunctions \citep{burrows2002}. These words are chosen subconsciously, and contain stable authorial habits \citep{kestemont2014}; the measure has proven robust across repeated evaluations and refinements \citep{hoover2004, argamon2008, evert2017}. 
This emphasis on style marks an important departure from much political-science text-as-data research. When the objective is to recover topics, positions, sentiment, frames, or policy content, frequent words are commonly removed as stopwords or downweighted through schemes such as TF–IDF so that rarer, more substantively distinctive terms dominate the representation. Stylometry reverses that priority. It focuses on the less directly content-bearing layer of language because frequent grammatical choices are ubiquitous, comparatively weakly governed by topic, and difficult for writers to regulate consistently. Function words may reveal little about a document’s subject, but their distribution can capture how a writer habitually constructs sentences and links clauses. Burrows’ Delta does not require dimensionality reduction: it directly yields distances between each text and the relevant authorial or group reference profiles in a single dimension.
The main text therefore reports the two most central stylometric diagnostics: character 3-gram UMAP plots and Burrows' Delta results. Where authorship or production labels exist---CRS analysts, ombudsman rapporteurs, tweet device, teleprompter use---we evaluate how well the projection corresponds to the labels, summarising cluster quality with mean silhouette widths over the labelled groups, computed on the two-dimensional UMAP coordinates \citep{rousseeuw1987}. In the two U.S. validation cases with two labels, we additionally report two-sample $t$-tests on the UMAP coordinates and the Burrows' Delta scores. Where no labels are available, as in the Orb\'an corpus, the exercise is fully inductive.

Every case study is accompanied by a discriminant-validity battery, reported in Appendix~D: the identical UMAP layout is re-colored by sentiment, text length, mean sentence length, mean word length, passive-voice density, and the coordination/subordination ratio, and the correlation of each factor with each UMAP dimension is computed. Low correlations indicate that the projection is not simply re-discovering document length or register. Dictionary-based sentiment scores, raw document length, sentence length, and related descriptive measures serve only as supplementary diagnostics and are reported in Appendix~D.

All clustering, dimensionality reduction, inferential statistics, and visualisation are carried out in \textsf{R}.

\section{Use Case 1: CRS Reports (American Law, Short-Report Series)}

\subsection{Data collection}

Congressional Research Service (CRS) reports are confidential, non-partisan policy documents produced by named staff analysts for Members of Congress. The corpus was scraped from the EveryCRSReport project,\footnote{\url{https://www.everycrsreport.com/}, accessed 10 July 2026.} which systematises and republishes these documents. To hold subject matter and institutional register approximately constant, we restrict the sample to reports classified by EveryCRSReport under a single topic label.\footnote{We choose EveryCRSReport's \emph{American Law} label over official CRSReports.gov public subject headings because the latter change over time. EveryCRSReport's backward-compatible topic areas are generated by grouping authors' specialisations extracted with regular expressions, including phrases such as \texttt{specialist in ...}, and by incorporating title and summary terms. The EveryCRSReport GitHub page provides the category source code; the \emph{American Law} display category includes cues such as ``American Public Law'', ``American National Government'', ``A Legal Analysis'', ``National Government'', ``Government Organization and Management'', and ``Legal Overview''.} Reports credited to two or more authors, as well as those without named authors that constitute roughly half of the corpus, are excluded from the analysis.

We test the stylometric methods on the American Law topic subset---1,109 single-authored reports by 282 distinct analysts after cleaning. Topic, audience, and a highly formal legal-prose register are therefore similar across every document, yet each report is attributed to a specific analyst.
CRS reports are issued in several standardized formats distinguished by their report-number prefix: long reports (\texttt{R}, \texttt{RL}), short reports (\texttt{RS}), and rapid-response Insight pieces (\texttt{IN}), among others. Because format dictates document length and template, it is itself a stylometric confound: two analysts may appear to differ simply because one writes long reports and the other short ones. We therefore begin with a single format family---the short-report series (\texttt{RS}).
The short-report series comprises 167 documents; we color the seven most prolific analysts and leave the remainder as grey background points.

\subsection{Results}

\begin{figure}[hbt!]	
  \centering
  \includegraphics[width=0.85\linewidth]{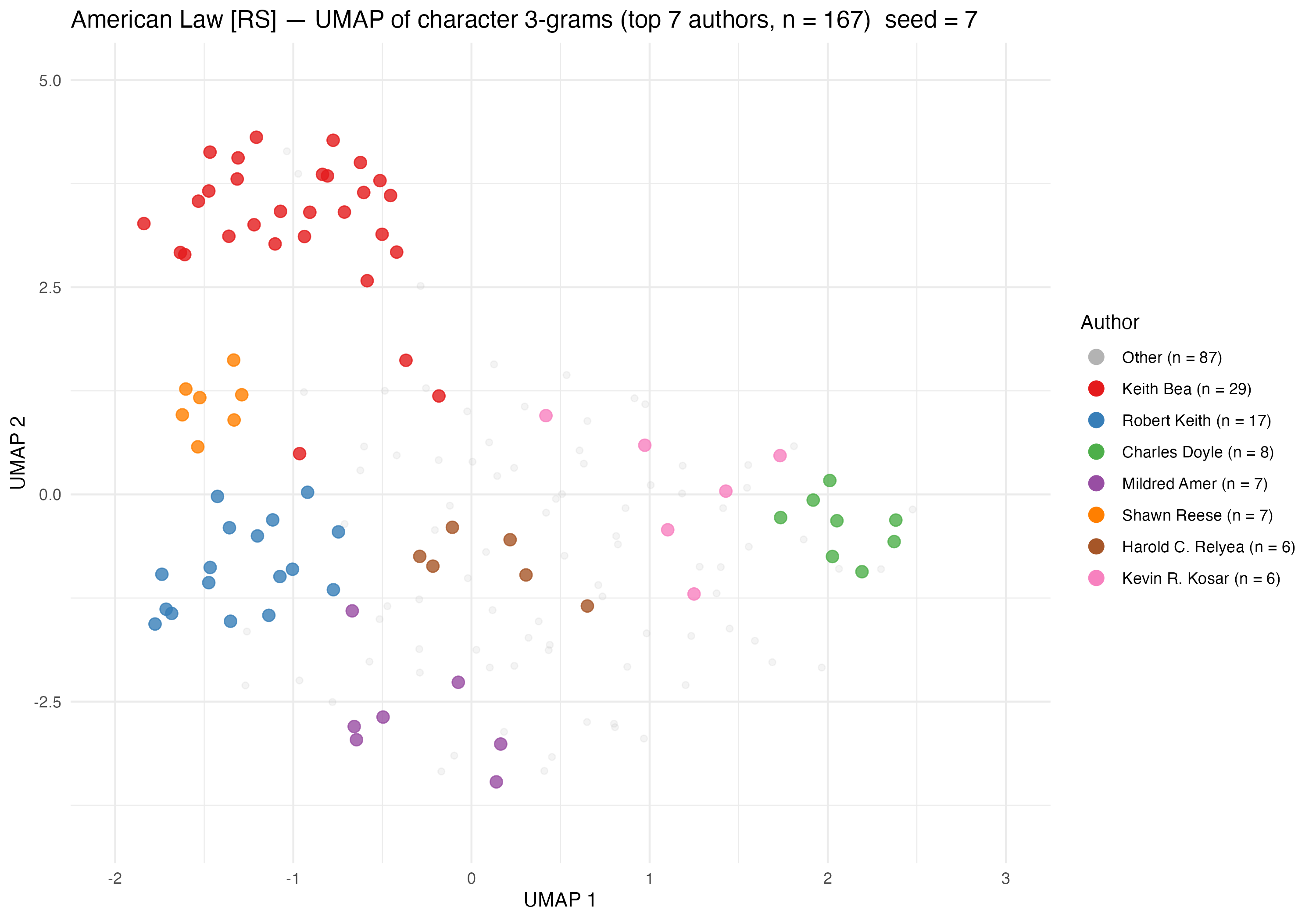}
  \caption{UMAP of character 3-gram features for the seven most prolific analysts in the CRS American Law short-report (\texttt{RS}) series ($n = 167$). Light grey = all other analysts. Holding document format constant alongside topic and register yields near-disjoint analyst clusters (silhouette $+0.47$).}
  \label{fig:crs_rs_umap}
\end{figure}

The character 3-gram UMAP (Figure~\ref{fig:crs_rs_umap}) yields the cleanest analyst separation in the entire study (mean silhouette width $+0.47$ across the colored analysts, computed on the plotted two-dimensional UMAP coordinates).
The three most prolific analysts resolve into three essentially disjoint islands: Keith Bea (29 reports) occupies a tight cloud in the upper-left, Robert Keith (17) a clearly separated cloud in the lower-left, and Charles Doyle (8) a distinct region on the right, with effectively no overlap among them.
The four less prolific analysts (Shawn Reese, Mildred Amer, Harold C. Relyea, and Kevin R. Kosar, each with six or seven reports) occupy their own local neighborhoods---more diffuse, but still internally coherent. Burrows' Delta corroborates the projection: within-analyst frequent-word profiles are markedly tighter than cross-analyst profiles (Appendix~B, Figure~A1).
This slice fixes topic area, audience, genre, and document format by construction, but one confound survives the design: in our sample, each analyst's reports concentrate in a distinct substantive area---27 of Bea's 29 short reports treat emergency-management and disaster statutes, 16 of Keith's 17 the congressional budget process, and 7 of Doyle's 8 federal criminal law---so analyst identity and subject matter are empirically entangled. Three checks nevertheless separate the two. The Burrows' Delta corroboration above rests on the 100 most frequent, and hence topically nearly empty, words; re-colouring the pooled American Law projection of Use Case 2 by official CRS topic tags yields a silhouette of $-0.31$, against $+0.41$ for analyst labels in identical coordinates; and analysts who write in several topic areas remain stylometrically themselves---in the wider CRS corpus, a report is more similar to its own author's reports on \emph{other} topics (mean trigram cosine $0.83$) than to other analysts' reports on the \emph{same} topic ($0.72$; Appendix~D.3, Table~A1). Given the three checks, the clusters track individual style, even though each analyst has a specialization.

\section{Use Case 2: CRS Reports (Full American Law Subset)}

\subsection{Data collection}

We next relax the format control of Use Case 1 and pool every report format within the American Law subset---long reports, short reports, Insight pieces, and legacy documents alike---yielding the full set of 1,109 documents attributed across 282 analysts.
The corpus now mixes document lengths and templates, reintroducing format as a potential secondary axis of stylistic variation. This tests whether the analyst signal isolated in Use Case 1 survives when the format is no longer held fixed. The eight most prolific analysts are colored.

\subsection{Results}

\begin{figure}[hbt!]
  \centering
  \includegraphics[width=0.85\linewidth]{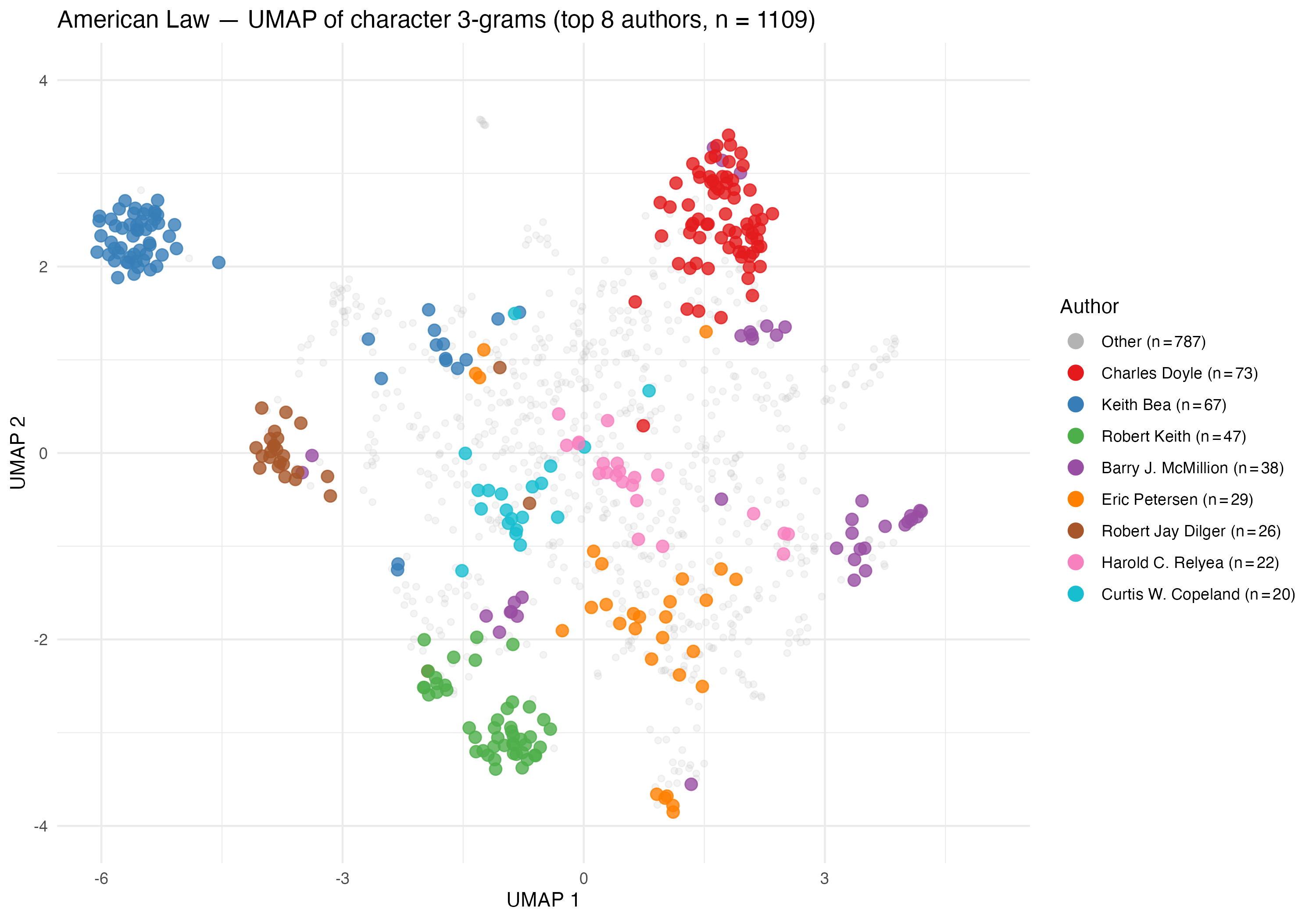}
  \caption{UMAP of character 3-gram features for the top eight analysts in the full CRS American Law subset ($n = 1{,}109$, all report formats pooled). Light grey = all other analysts. Analyst clusters remain clearly separated (silhouette $+0.41$) despite mixing document formats, but several analysts now occupy multiple islands, reflecting format-driven within-author splits.}
  \label{fig:crs_umap}
\end{figure}

Even with formats pooled, analyst-level structure remains strong, only marginally below the format-controlled short-report subset.
The most prolific analysts again resolve into tight, well-separated islands (Figure~\ref{fig:crs_umap}): Keith Bea (67 reports) in the upper-left, Charles Doyle (73) in the upper-right, Robert Keith (47) at the bottom, and Robert Jay Dilger (26) on the left.
The interior is visibly more mixed than in the short-report case, a second source of within-analyst spread introduced by format heterogeneity.
Burrows' Delta again shows within-analyst frequent-word profiles to be tighter than cross-analyst profiles (Appendix~B, Figure~A2).
That the silhouette barely declines ($+0.47 \rightarrow +0.41$) when document format is no longer held constant indicates that the authorship signal dominates the format signal: pooling four document templates only slightly blurs the analyst clusters.
A direct control rules out subject specialisation as the driver: re-colouring the same projection by the official CRS topic tags yields a silhouette of $-0.31$, against $+0.41$ for analyst labels in identical coordinates (Appendix~D.3, Figure~A5).

The pooled projection also reveals a pattern with methodological consequences: several analysts now occupy more than one island. Keith Bea's reports split into the large, dense cloud on the far upper-left and a smaller satellite group near the centre of the map; Robert Keith's documents form one dominant cluster at the bottom with a separate clump above and to its left; Barry J. McMillion and Eric Petersen likewise fragment into a principal cluster plus outlying groups. Because Use Case 1 showed that these same analysts form single coherent clusters once document format is held fixed, the most plausible reading is that the satellites are production artefacts: an analyst's \texttt{RS} short reports and \texttt{R}/\texttt{RL} long reports differ enough in template, length, and boilerplate that character trigrams register them as distinct stylistic products. Two further mechanisms plausibly contribute. CRS document templates were revised repeatedly over the decades the corpus spans, so an analyst's early and late reports inherit different boilerplate; and analysts change specializations over their careers, so their legal vocabulary changes too. Inductive stylometry recovers authorship within a production context: when format or period varies within a corpus, one writer's documents may form several clusters. The islands remain internally pure: format separates documents by the same author without merging documents from different authors. An individual cluster can therefore still support an authorship interpretation, although the mapping from clusters to authors may be many-to-one. But this phenomenon is familiar with text analysts, such as topic modelers, where not keeping the text context and format adequately constant often yields topic clusters correlated with such contexts and formats. Clearly similar care must be taken with stylometric assessments. 

Without analyst labels, the projection would still contain distinct islands and regions requiring substantive interpretation. The known analyst labels are what allow us to validate those regions as primarily authorial rather than merely exploratory clusters. This matters because it empowers research questions, such as attributing the roughly half of all anonymous CRS reports to their authors or finding and analyzing authorship patterns in fully anonymous political or policy documents. 

\section{Use Case 3: Hungarian Ombudsman Reports}

\subsection{Data collection}

Having established that the approach recovers individual hands in formal English legal prose, we next ask whether it travels to a typologically distant language. The Hungarian ombudsman publishes detailed reports on individual complaints and own-initiative investigations. We analyse the reports issued during M\'at\'e Szab\'o's tenure as Hungary's general ombudsman---Parliamentary Commissioner for Civil Rights from 2007 and, after the office was reorganised under the new constitution, Commissioner for Fundamental Rights from 2012 until 2013. Although every report is issued under the commissioner's name, each names the staff rapporteur (\emph{el\H{o}ad\'o}) who investigated the case and drafted the text, making this a labeled multi-author corpus of administrative legal documents in Hungarian.
The full corpus contains 886 reports spanning 76 rapporteur labels; after filtering the unattributed category, 75 named rapporteurs remain.
All documents share the same institutional genre, formal register, and legal subject matter, providing a tough cross-linguistic benchmark in which topic- and genre-driven clustering cannot account for rapporteur-level separation.

\subsection{Results}

\begin{figure}[!htb]
  \centering
  \includegraphics[width=0.85\linewidth]{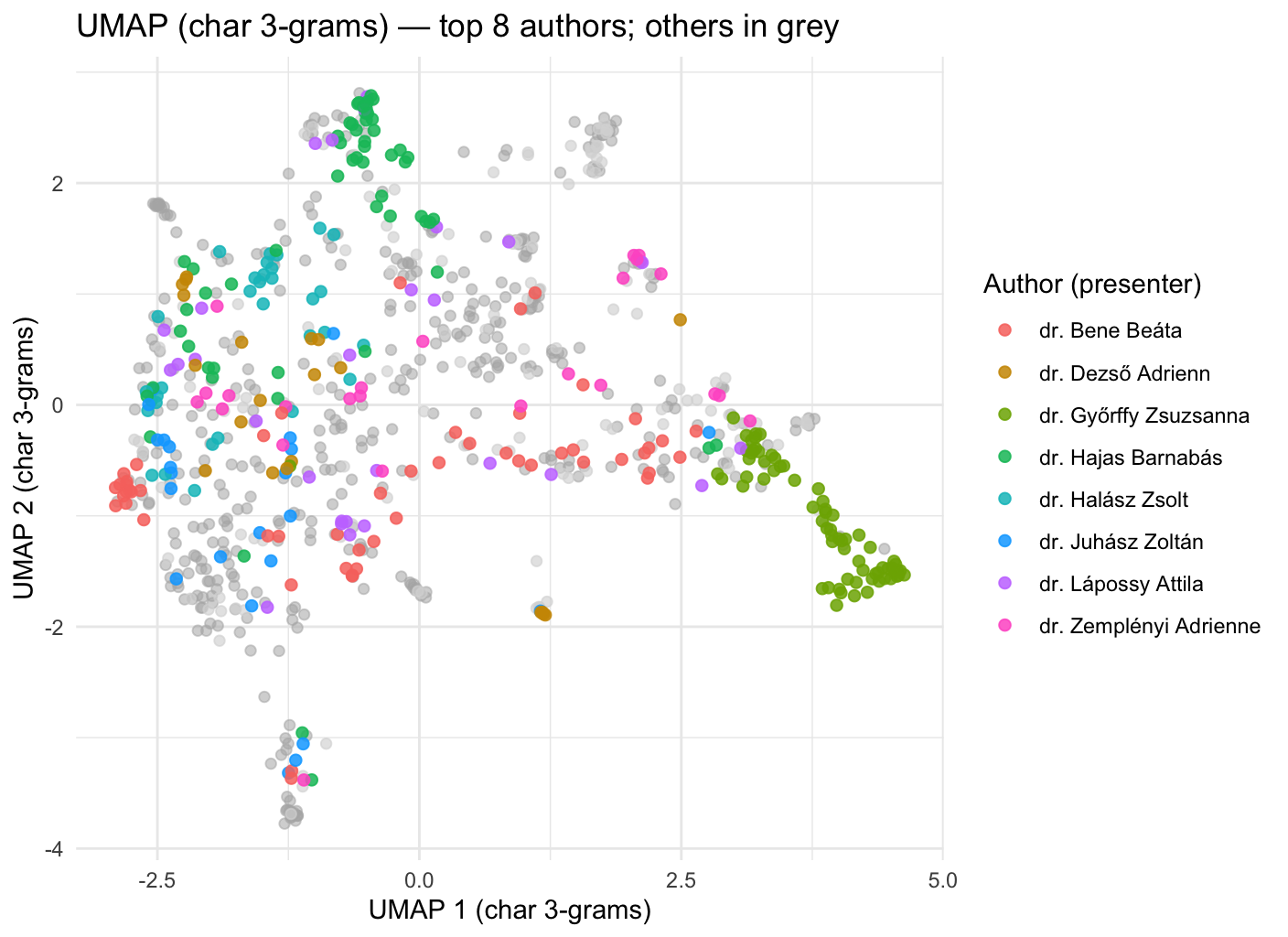}
  \caption{UMAP of character 3-gram features for the eight most prolific rapporteurs in the Hungarian ombudsman corpus (Szab\'o era, $n = 886$). Grey = all other rapporteurs. Several rapporteurs form clean islands (most visibly dr.\ Gy\H{o}rffy on the right); others remain diffuse.}
  \label{fig:ombudsman_3grams}
\end{figure}

The character 3-gram UMAP (Figure~\ref{fig:ombudsman_3grams}) reveals clear but uneven rapporteur-level structure among the eight most prolific drafters; the remaining rapporteurs appear as grey background points.
Several rapporteurs resolve into unmistakable islands: dr.\ Gy\H{o}rffy Zsuzsanna's reports form an elongated, nearly pure cluster along the right edge of the projection, dr.\ Zempl\'enyi Adrienne and dr.\ Hal\'asz Zsolt each occupy compact local neighborhoods, and dr.\ Hajas Barnab\'as and dr.\ Bene Be\'ata concentrate in identifiable regions of the map. Other rapporteurs remain diffuse, indicating that the strength of the recoverable fingerprint varies across individuals.
Because all documents are administrative Hungarian legal texts sharing genre, register, and broad subject matter, the separation that does emerge cannot be attributed to topic, genre, or language-register differences.
The full discriminant-validity battery in Appendix~D.4 (Figures~A6--A7) returns near-zero correlations for sentiment, length, sentence length, word length, and syntactic factors as well, the cleanest such result in the study.
The case demonstrates that character trigrams carry authorial signal even in an agglutinative, morphologically rich language, provided the underlying texts are written, formal, and independently drafted. It also previews the paper's central substantive point: every one of these reports went out under a single nominal author---the commissioner---yet the stylistic fingerprints that the method recovers belong to the staff who actually drafted them.

\section{Use Case 4: Trump Tweets by Device}

\subsection{Data collection}

Moving from long-form reports to the opposite extreme of document length, we examine the canonical example of device-specific tweets. During the 2016 campaign, Donald Trump's Twitter account was accessed primarily from an Android and an iPhone handset. Prior analysis established that the Android phone was usually in Trump's own hands, whereas the iPhone was staffed \citep{robinson2016}; subsequent linguistic work documented systematic register variation on the account over time \citep{clarke2019}. We downloaded the Trump Twitter Archive \citep{trumptwitterarchive} and define the analysis window from Trump's campaign launch on 16 June 2015 through Election Day on 8 November 2016. We exclude tweets sent from non-phone clients, rows the archive marks as retweets or deleted, and tweets that reproduce another account's words verbatim. The resulting corpus contains 4,000 composed phone tweets---1,866 from Android and 2,134 from iPhone.
\subsection{Results}

\begin{figure}[!htb]
  \centering
  \includegraphics[width=\linewidth]{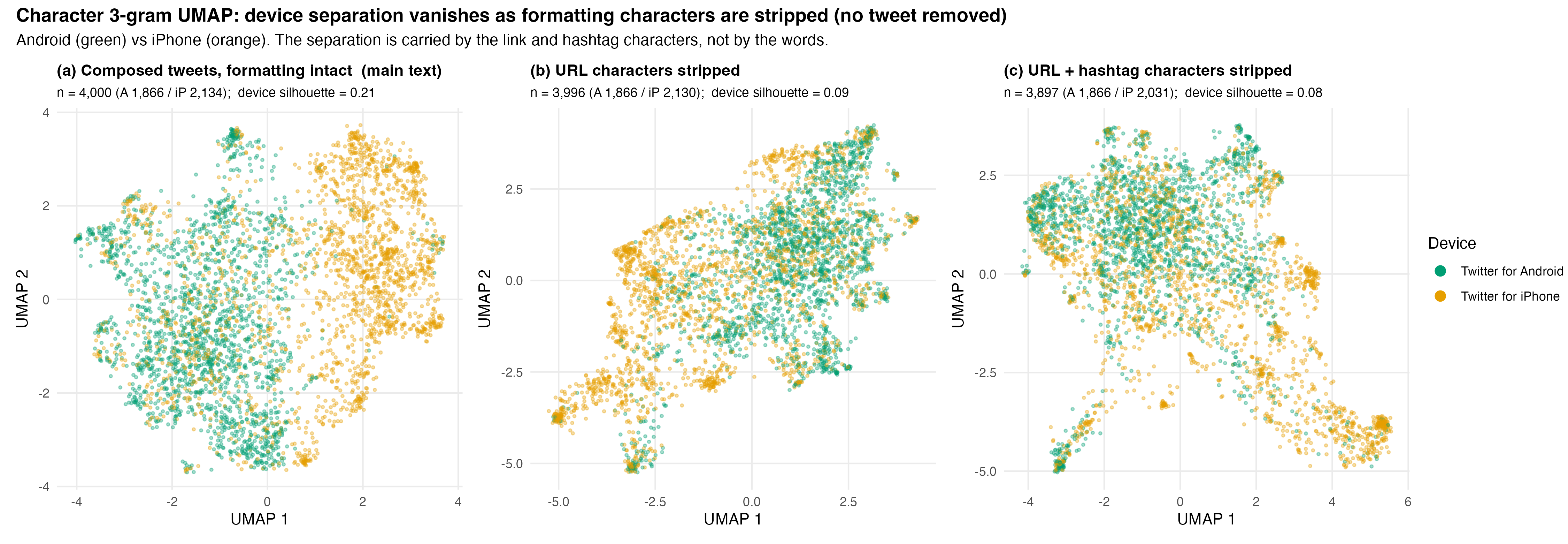}
  \caption{Character 3-gram UMAPs of the composed Trump campaign tweets
  ($n=4{,}000$), coloured by device (Android green, iPhone orange), under three
  formatting specifications. (a) With formatting intact, iPhone link-, hashtag-,
  and promotion-heavy tweets lie mostly on one side of the projection and Android
  unformatted prose on the other (mean device silhouette $0.21$). (b) Excluding
  URL characters reduces the silhouette to $0.09$. (c) Excluding hashtag characters
  as well reduces it to $0.08$.}
  \label{fig:tweet_umap_3grams}
\end{figure}

The character 3-gram UMAP differs significantly by device on both dimensions, although the groups overlap substantially (Figure~\ref{fig:tweet_umap_3grams}(a), Table~\ref{tab:tweet_stats_full}). iPhone tweets cluster in a link- and hashtag-heavy campaign register, whereas Android tweets are dominated by unformatted prose, including policy and media attacks, reactions, and updates. This pattern is more consistent with a difference in production register than with a clean division of authorship.

\begin{figure}[!htb]
  \centering
  \includegraphics[width=0.85\linewidth]{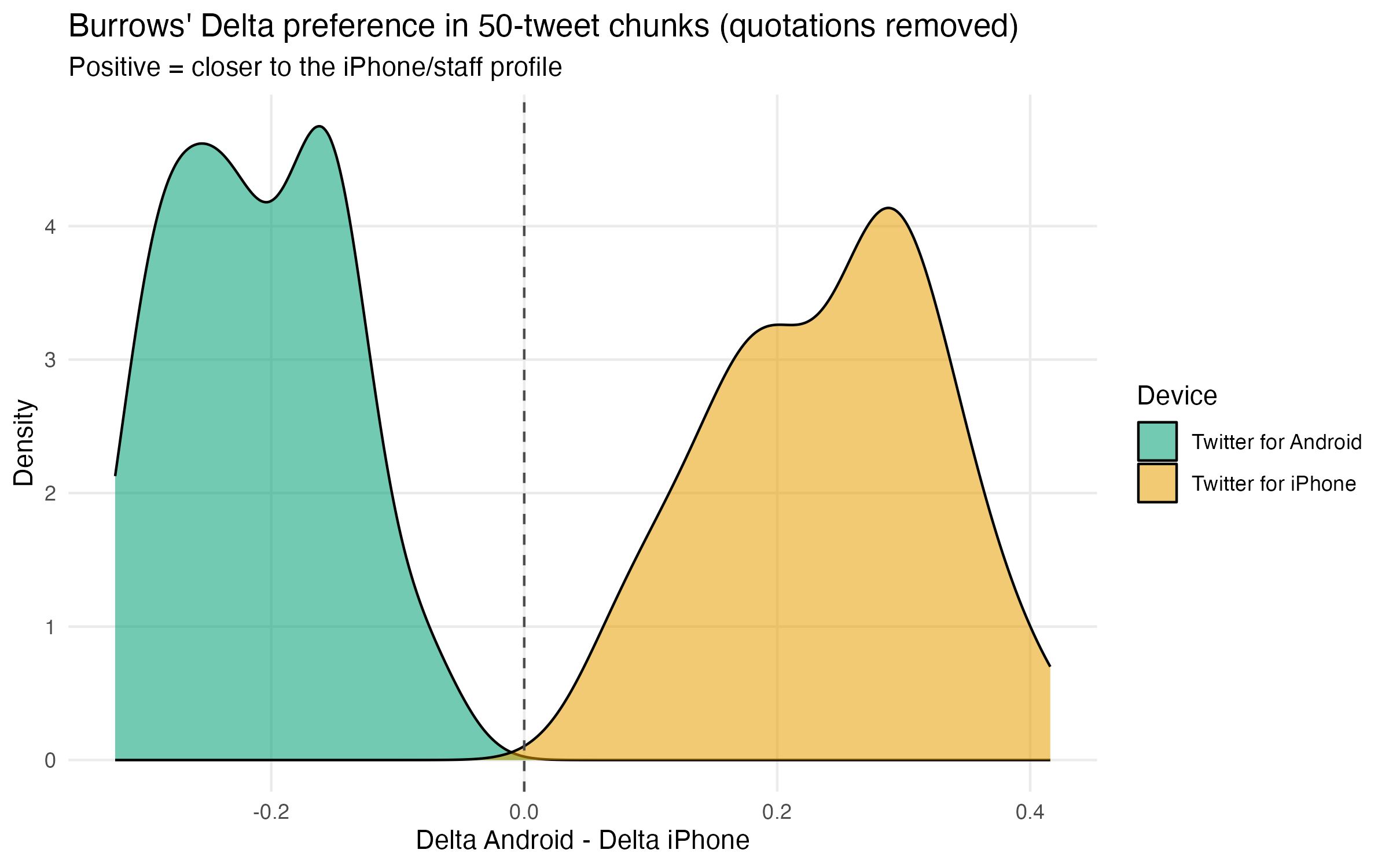}
  \caption{Burrows' Delta preference scores ($\Delta_{\text{Android}}-\Delta_{\text{iPhone}}$) for Android (green) and iPhone (orange) tweets, computed on chronological 50-tweet chunks within device. Positive values indicate greater similarity to the iPhone/staff profile. Aggregation renders the two device profiles almost perfectly separable.}
  \label{fig:burrow_tweets}
\end{figure}

The dictionary and frequent-word measures show the same production difference. iPhone tweets are shorter and more positive than Android tweets, although both gaps narrow after promotional formatting is excluded (Online Appendix), suggesting that they are partly associated with post type. Burrows' Delta also differs significantly by device.\footnote{Each device profile is the mean frequent-word vector for that device, so a tweet contributes marginally to its own reference profile; recomputing every score against leave-one-out profiles leaves the device ranking essentially unchanged (Online Appendix), so the separation is not an artefact of self-inclusion.} Because individual tweets are much shorter than the texts for which Delta was developed \citep{burrows2002, eder2015}, we calculate it both per tweet and on chronological 50-tweet chunks. The aggregated profiles are almost perfectly separable (Figure~\ref{fig:burrow_tweets}). The two streams are not contemporaneous: iPhone tweets occur later in the campaign, and their share of the account rises from two percent at launch to more than seventy percent by Election Day. Some device differences may therefore reflect the account's professionalisation over time \citep{clarke2019}.

Studies that use device as an authorship instrument therefore capture a meaningful but noisy division of labour on the account.

Formatting accounts for most of the device separation in the character 3-gram projection. Across the three embeddings in Figure~\ref{fig:tweet_umap_3grams}(a)--(c), the mean device silhouette is $0.21$ with formatting intact, $0.09$ without URL characters, and $0.08$ when hashtag characters are also excluded. The compact clusters that remain correspond mainly to repeated content, especially the campaign's \emph{Make America Great Again} sign-off, which appears in near-identical wording from both handsets. The small residual device difference follows the promotional register: iPhone staff posts remain shorter and more positive after links are excluded, while much of their prose uses Trump's own idiom, including first-person attacks, signature epithets, and self-promotion \citep{clarke2019}. Device is therefore a noisy proxy for authorship. Burrows' Delta still distinguishes the device labels, but that contrast cannot be interpreted as a clean difference between authors.

\begin{table}[ht]
\centering
\small
\begin{threeparttable}
\caption{Android vs.~iPhone tweets ($n=4{,}000$): dictionary metrics, Burrows' Delta, and character 3-gram UMAP dimensions}
\label{tab:tweet_stats_full}
\begin{tabular}{@{}lrrc@{}}
\toprule
 & \multicolumn{2}{c}{\textbf{Device mean}} & \\
\cmidrule(lr){2-3}
\textbf{Metric} & \multicolumn{1}{c}{\textbf{Android}} & \multicolumn{1}{c}{\textbf{iPhone}} & \textbf{Sig.}\tnote{a} \\
\midrule
\textit{Dictionary features} & & & \\
\quad Sentiment score            & $0.15$   & $0.72$   & * \\
\quad Tweet length (characters)  & $121.1$  & $106.4$  & * \\
\addlinespace
\textit{Burrows' Delta preference}\tnote{b} & & & \\
\quad Per tweet                  & $0.050$  & $0.075$  & * \\
\quad Per 50-tweet chunk         & $-0.211$ & $0.239$  & * \\
\addlinespace
\textit{Character 3-gram UMAP}\tnote{c} & & & \\
\quad Dimension 1                & $1.68$   & $-1.47$  & * \\
\quad Dimension 2                & $-1.62$  & $1.42$   & * \\
\bottomrule
\end{tabular}
\begin{tablenotes}
\footnotesize
\item[a] Two-sample $t$-tests; $^{*}p<0.0001$.
\item[b] $\Delta_{\text{Android}}-\Delta_{\text{iPhone}}$; positive values indicate greater similarity to the iPhone/staff profile.
\item[c] Tests on UMAP coordinates are descriptive corroboration only: the coordinates carry no meaningful units and the embedding induces dependence across observations (see Methods).
\end{tablenotes}
\end{threeparttable}
\end{table}

\section{Use Case 5: Trump Speeches (Teleprompter vs.~Off-the-Cuff)}

\subsection{Data collection}

Speeches add a layer that neither reports nor tweets possess: oral delivery, which interposes the speaker between whoever drafted the text and the transcript we observe. We gathered transcripts of 23 Trump campaign speeches from 2016--2017 and manually labelled 14 of them as teleprompter-assisted, based on the visible presence of a teleprompter in a video recording of the event; the remaining nine were coded as off-the-cuff. The speech corpus and teleprompter coding draw on the same Team Populism data used in the Guardian's ``Teleprompter Test'' interactive on scripted and off-the-cuff Trump speeches \citep{smith2019teleprompter}. All delivery coding, including the identification of off-script passages used below, was completed from the video record before, and independently of, the stylometric analysis.

\subsection{Results}

\begin{figure}[!htb]
  \centering
  \includegraphics[width=0.8\linewidth]{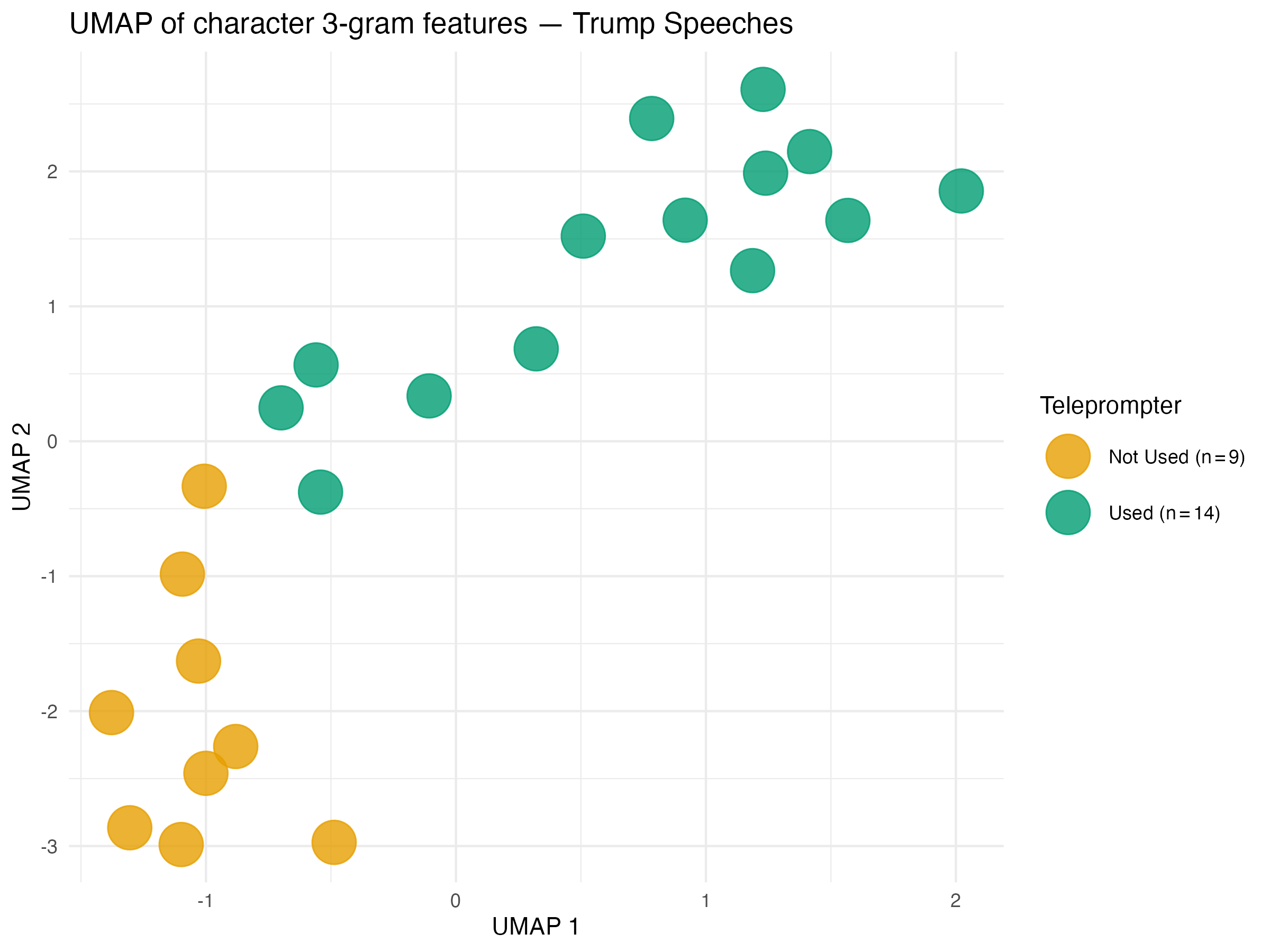}
  \caption{UMAP of character 3-gram features for 23 Trump campaign speeches (green = off-the-cuff, orange = teleprompted). The groups separate without overlap; the teleprompter speeches nearest the improvised cluster are those with documented off-script passages.}
  \label{fig:speech_umap_3gram}
\end{figure}

Table~\ref{tab:speech_stats_full} shows dictionary metrics, Burrows' Delta, and the 3-gram UMAP dimensions. Sentiment is the only dictionary feature with a significant gap---impromptu speeches are markedly more positive---while speech length does not differ between the groups. The character 3-gram projection (Figure~\ref{fig:speech_umap_3gram}) separates the impromptu and teleprompter speeches with no overlap, although the two groups are not spatially distant. The trigram--UMAP projection separates the delivery styles, whereas t-SNE and truncated-SVD projections of TF--IDF features (not shown) leave them overlapping.

\begin{figure}[!htb]
  \centering
  \includegraphics[width=0.8\linewidth]{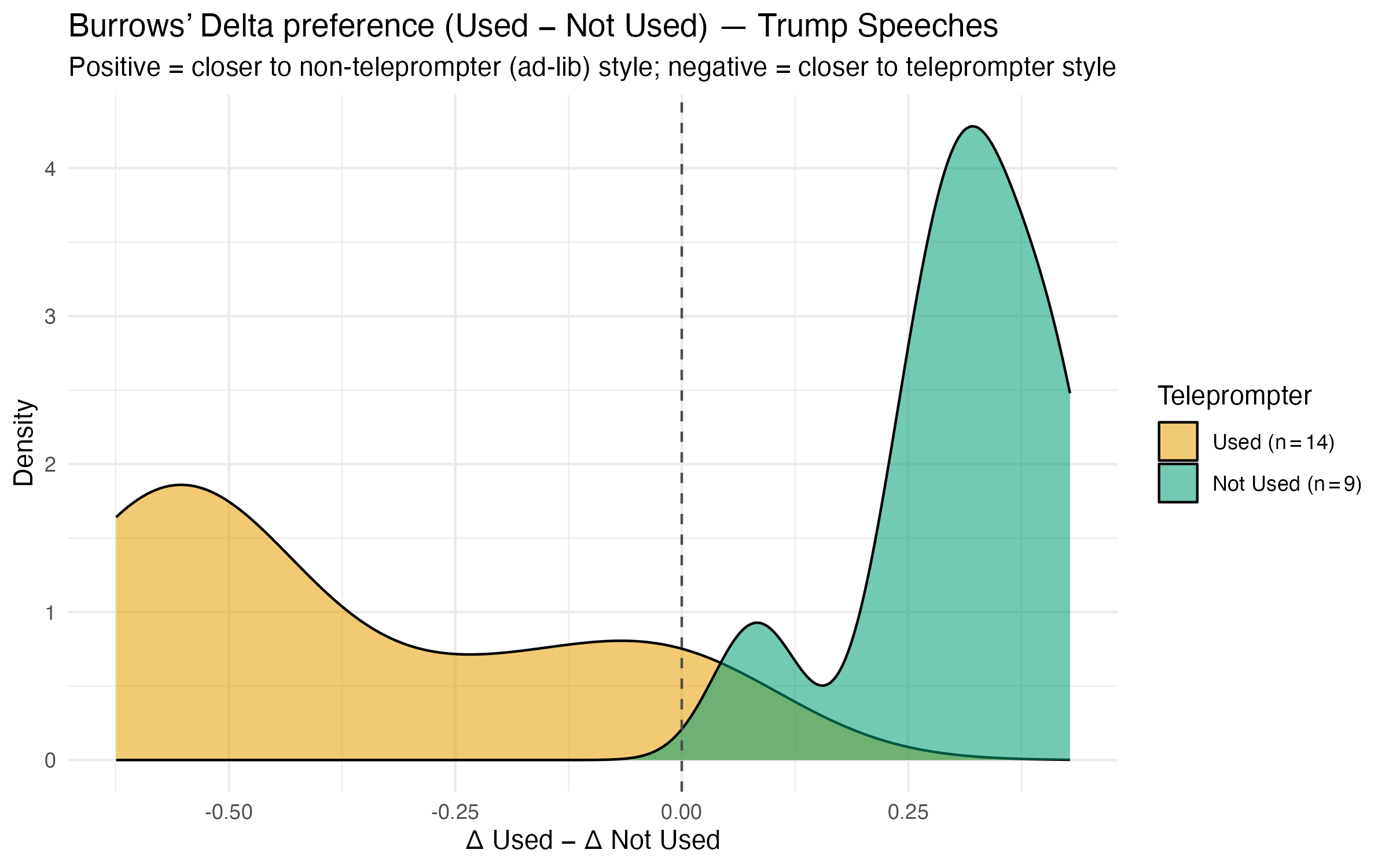}
  \caption{Burrows' Delta preference scores ($\Delta_{\text{Used}} - \Delta_{\text{NotUsed}}$; positive = closer to the improvised profile) for 23 Trump speeches (green = off-the-cuff, orange = teleprompted). The long tail of the scripted distribution toward zero reflects partially improvised teleprompter speeches.}
  \label{fig:trump_speeches_burrow}
\end{figure}

Positions within the scripted cluster correspond to observed departures from the prepared text. Teleprompter speeches extend from a compact, fully scripted core in the upper right toward the improvised cluster in the lower left. Video evidence shows repeated departures from the prepared text among speeches in the intermediate zone. The extreme case is the November 8, 2016 election-eve rally in Grand Rapids, Michigan: teleprompters were present, but the delivery was almost entirely improvised. Speeches with frequent, shorter ad-libs occupy intermediate positions, while addresses with almost no off-prompt passages, such as the August 20, 2016 rally in Fredericksburg, Virginia, lie at the scripted end. A speech-by-speech accounting of the observed off-script passages appears in Appendix~C. The projection therefore captures a gradient of scriptedness rather than a binary distinction: the farther a teleprompter speech lies from the scripted core, the more of its delivery was improvised. Burrows' Delta preference scores show a similar pattern (Figure~\ref{fig:trump_speeches_burrow}): scripted speeches have a long tail toward the improvised profile, although function words still separate the two delivery modes almost perfectly.

Despite the involvement of multiple speechwriters in a campaign, the scripted corpus contains no internal subclusters that could be attributed to individual writers. Several factors may account for this. Collaborative drafting and editing may homogenise individual styles, and the amount of text per writer may be too small for reliable attribution \citep{eder2015}. Delivery may also obscure individual styles: the transcripts record Trump's timing, repetitions, and interjections alongside the prepared text. The analysis distinguishes scripted from improvised delivery but does not identify individual speechwriters.

\begin{table}[ht]
\centering
\small
\begin{threeparttable}
\caption{Off-the-cuff vs.~teleprompted speeches: dictionary features, Burrows' Delta, and character 3-gram UMAP dimensions}
\label{tab:speech_stats_full}
\begin{tabular}{@{}lrrc@{}}
\toprule
 & \multicolumn{2}{c}{\textbf{Delivery-mode mean}} & \\
\cmidrule(lr){2-3}
\textbf{Metric} & \multicolumn{1}{c}{\textbf{Off-the-cuff}} & \multicolumn{1}{c}{\textbf{Teleprompted}} & \textbf{Sig.}\tnote{a} \\
\midrule
\textit{Dictionary features} & & & \\
\quad Sentiment score            & $108.0$    & $17.4$    & * \\
\quad Speech length (words)      & $17{,}868$ & $20{,}281$ & \\
\addlinespace
\textit{Burrows' Delta preference}\tnote{b} & & & \\
\quad Per speech                 & $0.28$     & $-0.32$   & * \\
\addlinespace
\textit{Character 3-gram UMAP}\tnote{c} & & & \\
\quad Dimension 1                & $-1.03$    & $0.66$    & * \\
\quad Dimension 2                & $-2.06$    & $1.32$    & * \\
\bottomrule
\end{tabular}
\begin{tablenotes}
\footnotesize
\item[a] Two-sample $t$-tests; $^{*}p<0.0001$. With 23 speeches we report significance levels rather than point $p$-values.
\item[b] $\Delta_{\text{Used}} - \Delta_{\text{NotUsed}}$; positive values indicate greater similarity to the off-the-cuff profile.
\item[c] Tests on UMAP coordinates are descriptive corroboration only: the coordinates carry no meaningful units and the embedding induces dependence across observations (see Methods).
\end{tablenotes}
\end{threeparttable}
\end{table}

\section{Use Case 6: Orb\'an's Speeches in Hungarian}

\subsection{Data collection}

The final case examines whether speeches attributed to Hungary's prime minister, Viktor Orb\'an, show evidence of multiple authorship under conditions that make validation especially difficult. First, unlike in the previous case studies, no authorship labels are available to validate any clusters. Second, the texts are orally delivered and collaboratively produced, two features shared with the Trump corpus, where no speechwriter clusters emerged. Third, Hungarian has roughly 13 million native speakers, far fewer than English. It belongs to the Finno-Ugric branch of the Uralic family and has highly agglutinative morphology that differs substantially from the grammatical structure of Indo-European languages.

The Hungarian government website (kormany.hu) posts the most recent speeches of the Prime Minister, while older ones can be retrieved by parsing the Wayback Machine, which caches the most important web pages. We scraped a corpus of approximately 1,200 texts attributed to Orb\'an (2016--2022, only about half of them unique). The corpus also included interviews, social-media posts, and press conferences, which could form genre-based clusters unrelated to authorship. We filtered the delivered speeches in three steps. First, we used a large language model to classify texts as speeches, interviews, or press conferences and removed roughly one-fifth of the corpus (the prompt and specifications appear in Appendix~A). Second, we applied dictionary-based filtering to eliminate chunks of English speech, social-media post announcements, and post-speech Q\&A sessions. This yielded 466 speeches, which we finally spot-checked manually.

\subsection{Results}

To explore stylistic structure, we project the high-dimensional 3-gram matrix with UMAP (cosine distance, 15 nearest neighbours, minimum distance 0.05). Because no authorship labels exist to validate against, we instead probe the layout with the full discriminant-validity battery, re-colouring the identical projection by six non-authorship factors that could plausibly organise it: dictionary-based affect (tokens joined to the \emph{poltextLAB} Hungarian political sentiment lexicon \citep{ring2024}, $+1$ for positive and $-1$ for negative matches, summed within document and length-normalised), text length, mean sentence length, mean word length, passive-voice density, and the coordination/subordination ratio. 
Across the hyperparameter settings examined, the projection contains no stable clusters (Appendix~D.7, Figure~A10). The six variables reveal no stable divisions; neither do comparisons between early and late speeches or domestic and international audiences. The limited structure in the projection is associated with sentence-level formality rather than authorship. Mean sentence length has the strongest correlation ($|r_{\text{U2}}| = 0.42$), followed by dictionary sentiment ($|r_{\text{U2}}| = 0.30$) and mean word length ($|r_{\text{U1}}| = 0.27$). Text length is nearly unrelated to either axis ($|r_{\text{U1}}| = |r_{\text{U2}}| = 0.05$), indicating that the projection does not simply separate brief ceremonial remarks from long programmatic keynotes. Passive voice and the coordination/subordination ratio are negligible. Absolute correlations are reported because the orientation of each UMAP axis is arbitrary. In short, the corpus contains too little recoverable stylistic structure for inductive stylometry in this setting.
The result resembles the Trump case in one respect: no speechwriter clusters emerge within the scripted texts. Unlike the Trump corpus, however, the Orb\'an corpus offers no scripted/improvised contrast against which to validate the projection. Three factors may contribute to this result. Agglutinative morphology spreads stylistic signal across an enormous trigram vocabulary, diluting the frequency profile of any individual habit. Oral delivery and transcription overwrite the drafters' orthographic fingerprints with the speaker's own cadence. And the institutional editing process of a prime ministerial speechwriting office harmonises whatever individual signal survives the first two filters. The case marks the boundary of classical frequency-based stylometry, and we return to what might lie beyond it in the discussion.

\FloatBarrier
\section{Discussion}

This paper demonstrates that a classical stylometric toolbox---character 3-grams and Burrows' Delta on frequent words, with UMAP for dimensionality reduction---recovers important authorship or production signals in five of six politically diverse corpora. The approach generalizes across text length (from tweets to long-form reports), language (English and Hungarian), and modality (written versus orally delivered), but not unconditionally. Comparing the successful cases with the unsuccessful case identifies the conditions under which the method works.

Four conditions appear to govern when inductive stylometry succeeds. First, morphology affects performance but does not by itself determine success: character trigrams perform best on morphologically simple English, while the ombudsman case shows that agglutinative Hungarian is not a barrier when texts are written, formal, and independently drafted. Second, institutional editing must not erase author-level differences. Our CRS and ombudsman cases succeed precisely because analysts and rapporteurs draft independently and are not subject to heavy cross-author revision; the speechwriter-level failures in both speech corpora are the mirror image of the same condition. Third, oral delivery places the speaker between the drafters and the transcript: delivery is a form of re-authorship. We can still tell scripted speech from improvised speech, but we can no longer tell the drafters apart. The single complete failure---Orb\'an's speeches---combines all three adverse conditions at once: agglutinative language, oral delivery, and heavy institutional editing. Fourth, individual texts must be long enough to provide stable feature coverage. Burrows' Delta was developed for texts upward of roughly 1,500 words \citep{burrows2002}, and attribution reliability degrades sharply in small samples \citep{eder2015}; the tweet corpus sits near the lower limit, and chunking into 50-tweet aggregates was required to stabilise the Delta profiles.

A UMAP projection cannot identify authors on its own. Apparent clusters may arise from topical, formatting, or production differences as well as authorship \citep{marx2024}. In the pooled CRS corpus, documents by the same analyst form separate format-based clusters, making the mapping from clusters to authors many-to-one. In the tweet corpus, the Android/iPhone separation reflects production register---link-and-promotion formatting versus composed prose---rather than two distinct authors. Each projection is therefore evaluated alongside Burrows' Delta on function words, which carry minimal topical signal \citep{kestemont2014, evert2017}, checks against available labels, and the discriminant-validity tests reported in Appendix~D. Interpretation rests on this supporting evidence, not on the projection alone.

Recovering authorship signals in institutional texts has direct consequences for how we evaluate political communication. If tweets, speeches, or policy documents bear the stylometric mark of a specific staffer, speechwriter, or institutional hand, claims about the nominal author's beliefs, intentions, or rhetoric risk misattribution. In the tweet corpus, the Android/iPhone split captures a real difference in account use but does not map cleanly onto authorship; stylometry helps identify where the proxy breaks down. The framework could also be used to study speechwriter turnover, changes in the authorship of institutional texts, and staff involvement in ostensibly personal political communication.

The Orb\'an failure also charts the path forward. Where frequency-based features run out of signal, learned authorship representations---neural embeddings trained specifically to encode who writes rather than what is written \citep{uar-emnlp2021}---offer a language-agnostic alternative that may survive translation, transcription, and editing better than surface trigrams. The six corpora provide benchmarks for testing these representations because the divisions recovered by classical methods are already documented. As large language models enter campaign and government drafting, the question ``whose line is it anyway?'' becomes harder to answer, increasing the need for validated authorship methods.

Political science usually treats the nominal speaker as the author of a political text. This paper shows that the assumption can be tested. We chose corpora where authorship or production labels was known in advance---analysts, rapporteurs, devices, teleprompters. Inductive stylometry has shown useful to identify the analysts behind institutional prose, separate a politician's own tweets from his staff's, and find the moments a speaker leaves the script. Because it is validated where authorship is known, it can be further applied where it is unknown: the unsigned half of the CRS corpus, anonymous official reports, ghost-written statements, etc. We also identified possible failure modes. When heavy institutional editing, oral delivery, and complex morphology come together, the individual style becomes more obscure. 
\section*{Data Availability Statement}

Replication materials---including all corpora and the \textsf{R} analysis code will be deposited in the Political Analysis Dataverse upon acceptance.

\section*{Supplementary Material}

The Appendix contains the LLM classification prompt used to filter the Orb\'an corpus (Appendix~A), Burrows' Delta density plots for the two CRS corpora (Appendix~B), the extended speech-by-speech analysis of off-script passages in the Trump teleprompter corpus (Appendix~C), and the full discriminant-validity battery for all six use cases (Appendix~D).

\FloatBarrier

\clearpage
\appendix
\setcounter{section}{0}
\setcounter{figure}{0}
\setcounter{table}{0}
\renewcommand{\thesection}{\Alph{section}}
\renewcommand{\thefigure}{A\arabic{figure}}
\renewcommand{\thetable}{A\arabic{table}}
\section*{Appendices}

\section{AI Prompt for Speech Classification (Orb\'{a}n Corpus)}

To filter the Orb\'{a}n corpus (Use Case 6), each text was classified via the following prompt submitted to the OpenAI ChatGPT o3 model:

\bigskip
\noindent\fbox{%
  \parbox{0.92\linewidth}{%
    \textbf{Prompt:} You are given a text snippet of a Hungarian speech of Viktor Orb\'{a}n.
    Classify it into exactly one of the following categories by returning ONLY the number:\\[4pt]
    1 = Interview\\
    2 = Press conference\\
    3 = Delivered speech\\[4pt]
    No extra words or punctuation, just the digit 1, 2, or 3.
  }%
}
\bigskip

Texts that received classification 3 were retained; classifications 1 and 2 were discarded.
The prompt was applied to text snippets of approximately 500 characters rather than full transcripts, both to reduce API cost and because brief excerpts are often sufficient to distinguish the conversational register of an interview from the monologic register of a delivered speech.
The remaining corpus was further cleaned with dictionary-based filters (removal of English-language chunks, social media post headers, and post-speech Q\&A passages appended to some transcripts).
This two-stage filtering yielded 466 unique, cleanly delimited speeches used in the main analysis.

To assess classification accuracy, a human reviewer manually checked a random sample of 50 classified texts; the human and LLM labels agreed in all 50 cases for the dominant speech-versus-interview distinction.
This result describes the reviewed sample and does not establish the corpus-wide misclassification rate. The classification was used as a practical data-preparation procedure rather than as a rigorously validated annotation protocol, and systematic errors may remain in edge cases (e.g., transcripts that begin mid-sentence, or speeches with unusually long Q\&A preambles).

\section{Burrows' Delta for the CRS Corpora (Use Cases 1--2)}
\label{sec:crs_delta}

The figures below report the Burrows' Delta evidence referenced in the main text for the two CRS corpora. They show, by analyst, the distribution of each document's Delta distance to its own author's frequent-word style profile. In both the format-controlled short-report series and the pooled American Law subset, within-analyst profiles are markedly tighter than cross-analyst profiles. This result is consistent with the trigram--UMAP projections and uses frequent function words, which carry little topical signal.

\begin{figure}[!htb]
  \centering
  \includegraphics[width=0.9\linewidth]{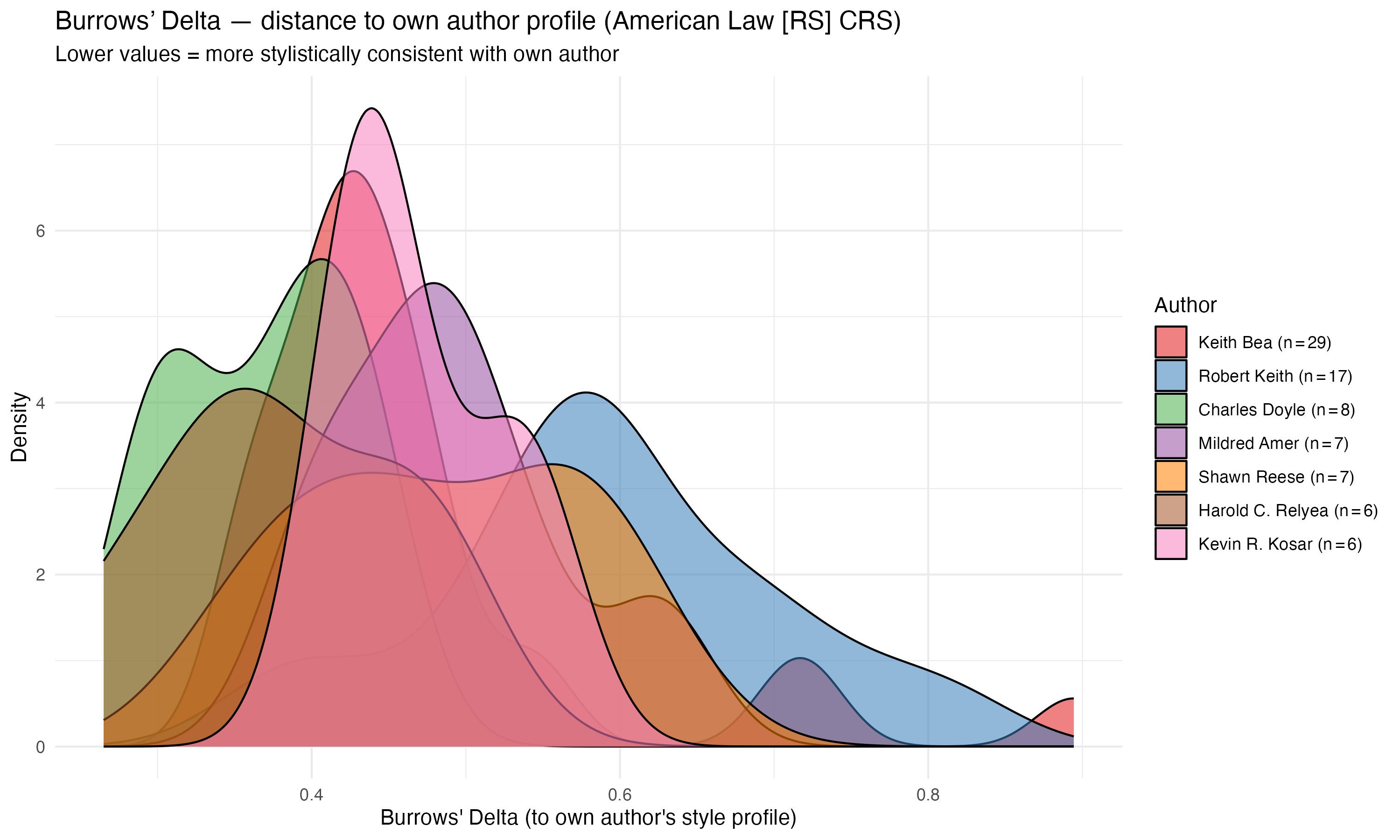}
  \caption{Burrows' Delta distance to each analyst's own style profile, CRS American Law short-report (\texttt{RS}) series (Use Case 1). Lower values indicate greater within-analyst stylistic consistency.}
  \label{fig:crs_rs_delta_app}
\end{figure}

\begin{figure}[!htb]
  \centering
  \includegraphics[width=0.9\linewidth]{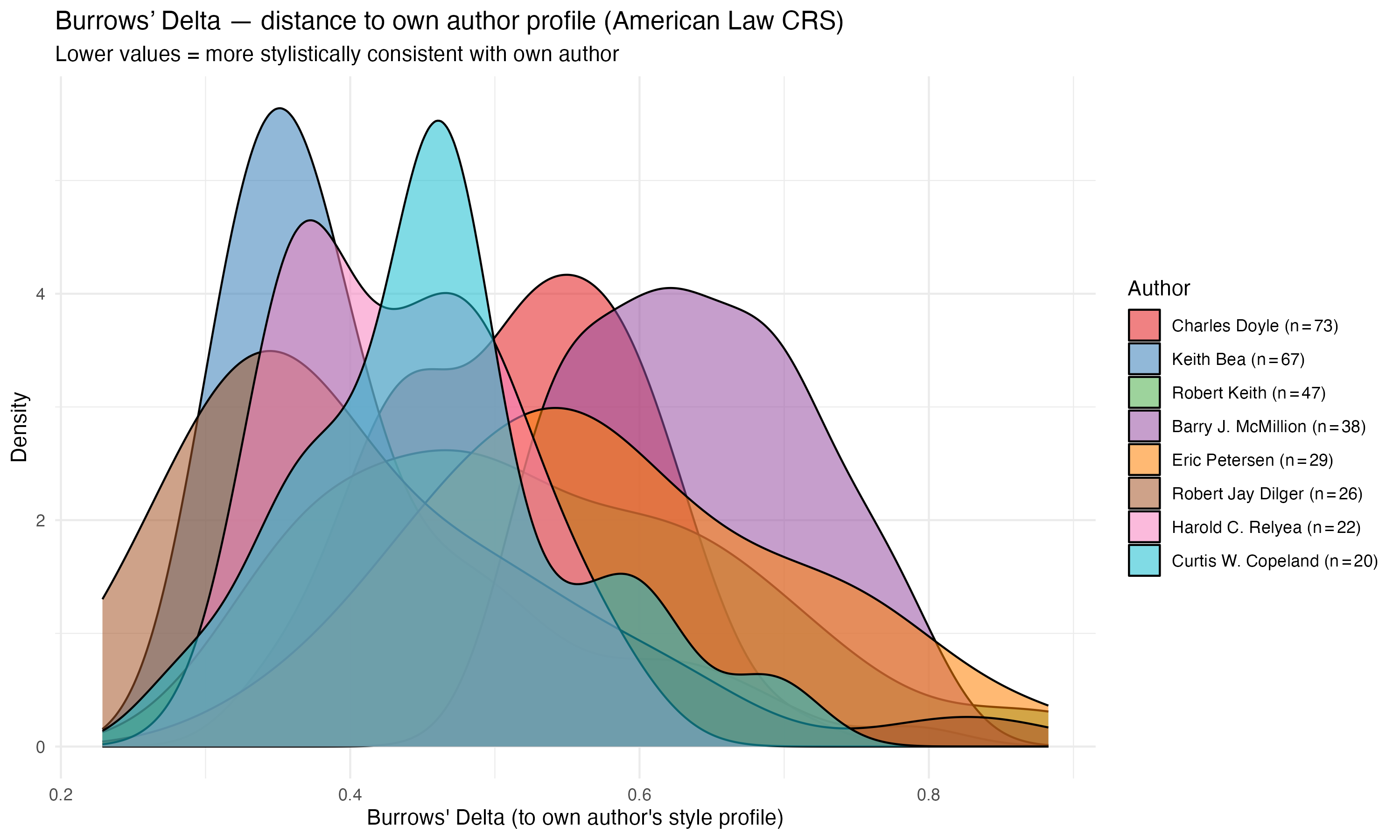}
  \caption{Burrows' Delta distance to each analyst's own style profile, full CRS American Law subset (Use Case 2). Lower values indicate greater within-analyst stylistic consistency.}
  \label{fig:crs_delta_app}
\end{figure}

\clearpage

\section{Use Case 5: Trump Speeches --- Extended Analysis of UMAP Structure}
\label{sec:speeches_extended}

The character 3-gram UMAP shows a strong overall separation between speeches delivered with and without a teleprompter. Most non-teleprompter speeches occupy the upper portion of the figure, while most teleprompter speeches form a separate group; several clearly scripted appearances lie at the lower-left extreme. This distribution associates teleprompter use with a different textual and stylistic profile.

The coding records the physical presence of a teleprompter; it does not show that Trump read continuously from prepared text. In several coded speeches, Trump appears to depart briefly from the script, making them more similar to impromptu speeches.

Teleprompters were present at the November 8, 2016 Final Election Eve Rally in Grand Rapids, Michigan, but the speech appears to have been delivered in an almost entirely improvised manner. The presence of the equipment therefore does not by itself indicate a fully scripted delivery style.

A second case is the October 13, 2016 West Palm Beach, Florida address, ``Crossroads of Our Nation,'' a teleprompter-coded speech in which the speaker repeatedly deviates from the prompt for short intervals.
Off-script passages appear around 7:00, 8:00, and 8:30, followed by more extended improvised stretches beginning around 15:00, including a segment from roughly 17:53 to 19:07.
Additional shorter departures occur around 19:26--19:40, 19:58, 20:09, 20:20, 20:48, 21:46, 23:00--23:15, 26:40, 31:20--31:43, 32:15--32:55, 37:47--37:55, 38:00--39:15, and 45:00.
Many of these moments appear to involve jokes, ad libs, or spontaneous elaborations. Some speeches coded as using a teleprompter therefore contain a nontrivial share of improvised speaking time.

Other teleprompter speeches adhere more closely to the prepared script. The June 7, 2016 Primary Victory Speech in Briarcliff Manor, New York appears to contain only occasional short departures, typically brief phrases or minor insertions. The August 20, 2016 rally in Fredericksburg, Virginia does not appear to contain meaningful off-prompt stretches and occupies the most distant part of the teleprompter cluster in the embedding.

The teleprompter category includes speeches that are strongly scripted throughout, speeches with short impromptu segments, and a few that may be largely improvised despite the presence of teleprompter hardware.

\section{Discriminant Validity: UMAP Colored by Non-Authorship Factors}
\label{sec:dv}

\subsection{Rationale and Method}

One alternative explanation for the observed UMAP clustering is that surface-level textual properties, including length, sentiment, and syntactic register, shape the projection. If so, the clusters need not reflect authorship-level stylometric signals. We examine this possibility by re-coloring the same character 3-gram UMAP projections used in the main text according to six factors:

\begin{enumerate}
  \item \textbf{Sentiment score} --- dictionary-based positive/negative balance (\texttt{bing} for English; \texttt{poltextLAB} for Hungarian).
  \item \textbf{Text length} --- total character count per document.
  \item \textbf{Mean sentence length} --- average number of words per sentence.
  \item \textbf{Mean word length} --- average number of characters per word.
  \item \textbf{Passive-voice density} --- passive constructions per word, identified by a rule-based tagger.
  \item \textbf{Coordination/subordination ratio} --- count of coordinating conjunctions divided by count of subordinating conjunctions.
\end{enumerate}

Each figure below reproduces the UMAP layout from the main text in a 3\,$\times$\,2 panel grid, replacing the authorship colour label with a continuous colour gradient for one factor per panel.
Pearson $r$ between each factor and each UMAP dimension is printed as a subtitle in every panel.
Low correlations and the absence of a systematic gradient matching the authorship clusters support a stylometric authorship interpretation and make these potential confounds less plausible explanations for the observed separation.
Where correlations are elevated, we discuss the substantive interpretation below.
The subsections follow the order of the use cases in the main text.

\subsection{Use Case 1: CRS Reports (American Law, Short-Report Series)}

\begin{figure}[!htb]
  \centering
  \includegraphics[width=\linewidth]{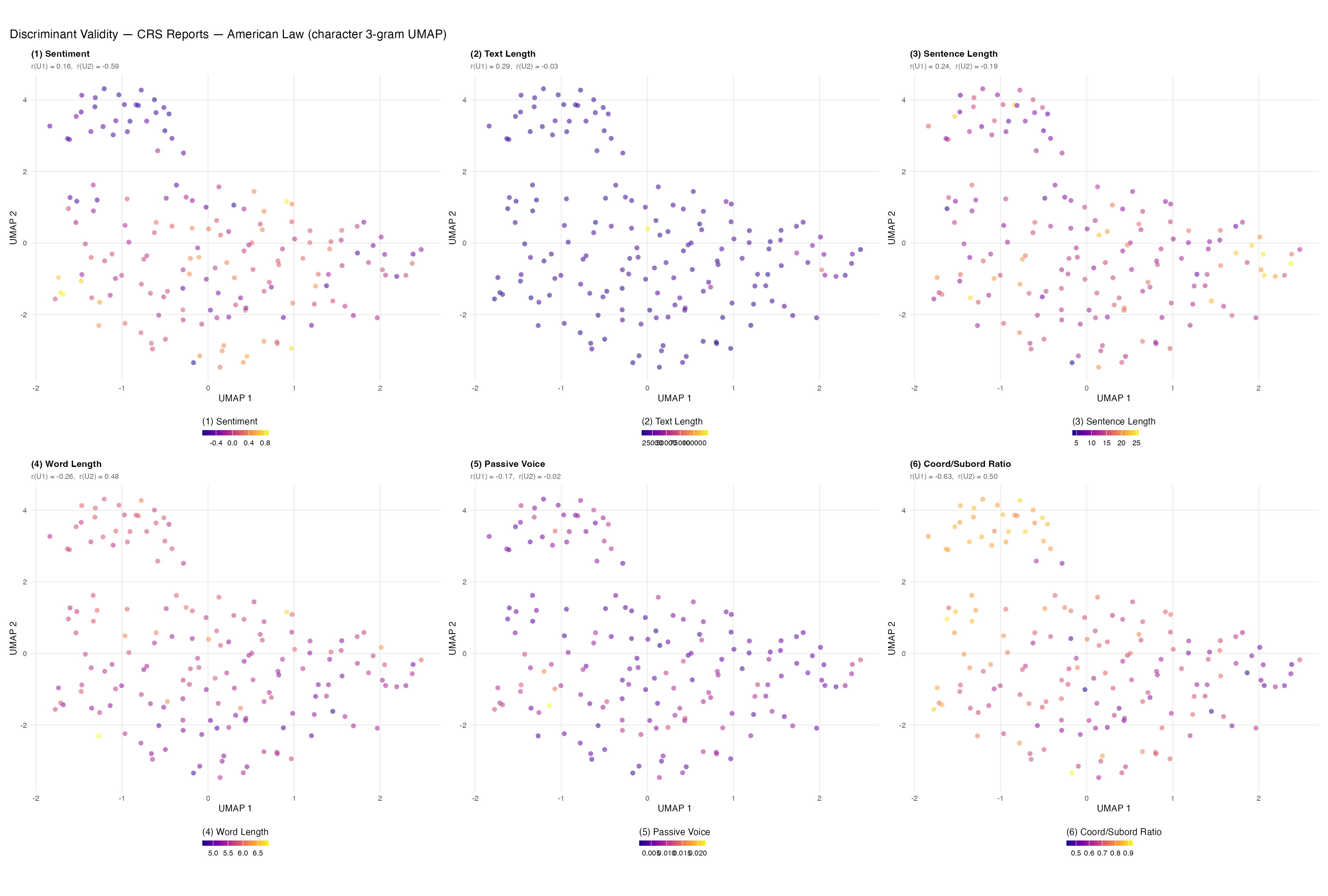}
  \caption{Discriminant validity for the CRS American Law short-report (\texttt{RS}) series (Use Case 1).
  The same character 3-gram UMAP as in the main text, re-coloured by six alternative factors.
  Pearson $r$ with each UMAP dimension is printed as a panel subtitle.}
  \label{fig:dv_crs_rs}
\end{figure}

\textbf{Interpretation.}
The short-report series is the most tightly controlled corpus in the study: topic area, institutional audience, formal genre, and document format are all held constant by construction.
Because corpus-level genre and format are fixed, an association between a surface factor and the projection axes would reflect systematic differences among analysts within this corpus.

\textit{Observed pattern.}
Text length is the most plausible remaining confound, but its correlations are weak ($r_{\text{U1}} = 0.29$, $r_{\text{U2}} = -0.03$). Mean sentence length ($r_{\text{U1}} = 0.24$, $r_{\text{U2}} = -0.19$) and passive-voice density ($r_{\text{U1}} = -0.17$, $r_{\text{U2}} = -0.02$) are also weak.
Three factors reach more substantial magnitudes: the coordination/subordination ratio ($r_{\text{U1}} = -0.63$, $r_{\text{U2}} = 0.50$), sentiment ($r_{\text{U2}} = -0.59$), and mean word length ($r_{\text{U2}} = 0.48$).
The gradients align with the analyst clusters. The clearest example is the Keith Bea cluster at the top of the projection, which has both a high coordination/subordination ratio and low dictionary sentiment.
Because every document shares format, topic, audience, and register, these associations are consistent with analyst-level differences in syntactic architecture (a preference for coordinate over subordinate constructions) and in vocabulary that the sentiment dictionary scores as affective. These may be author-consistent habits encoded by the character trigrams rather than corpus-level confounds.
Dictionary sentiment scores over specialised legal vocabulary should in any case be read cautiously, given the sparse coverage of the \texttt{bing} lexicon in this register.

\subsection{Use Case 2: CRS Reports (Full American Law Subset)}

\begin{figure}[!htb]
  \centering
  \includegraphics[width=\linewidth]{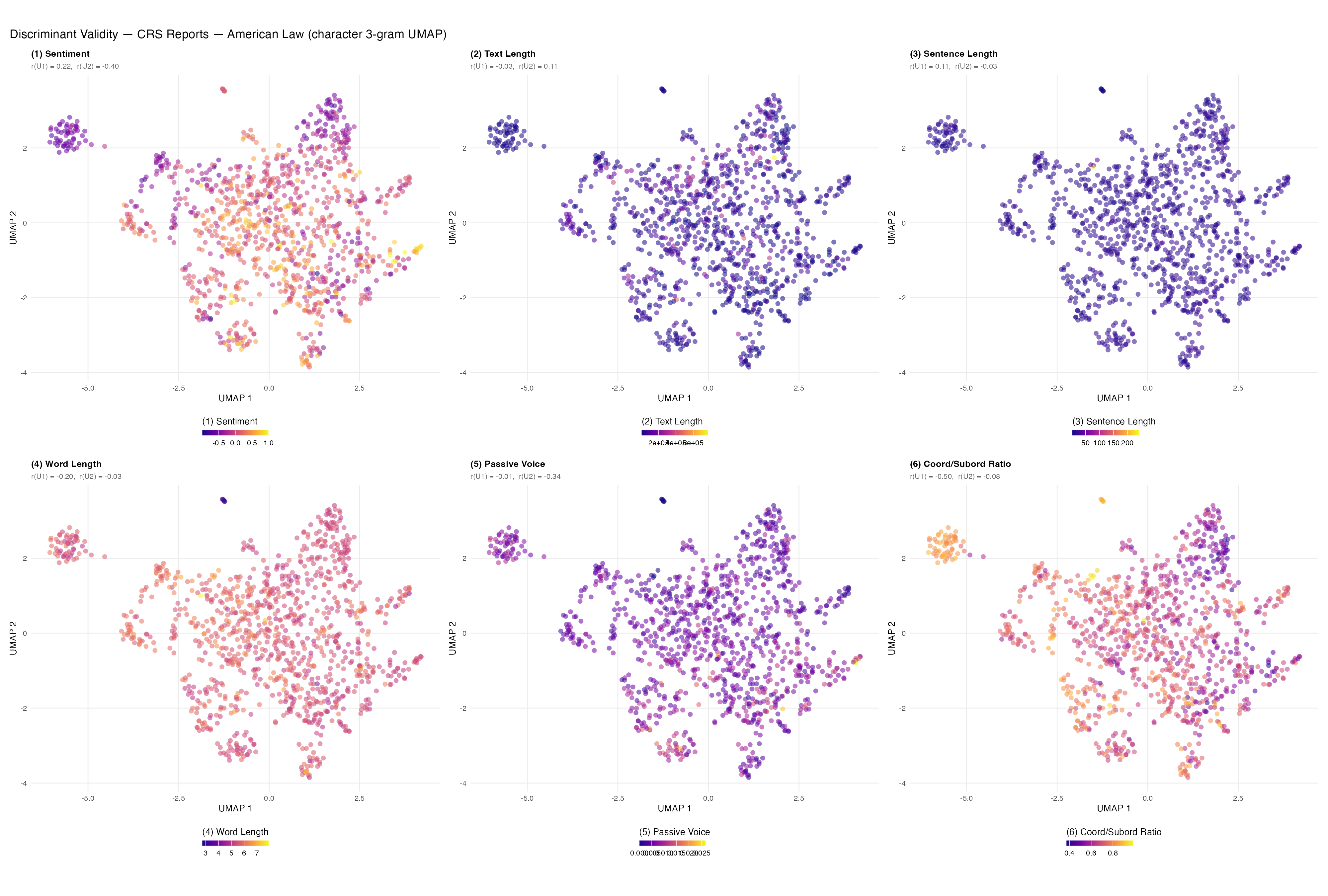}
  \caption{Discriminant validity for CRS Reports, full American Law subset (Use Case 2).
  Character 3-gram UMAP re-coloured by six alternative factors. One extreme point is omitted from this display to prevent compression of the main cluster; this display choice is separate from the outlier filter described in the main text.
  Pearson $r$ with each UMAP dimension is printed as a panel subtitle.}
  \label{fig:dv_crs}
\end{figure}

\textbf{Interpretation.}
Every document in the CRS American Law corpus shares the same topic area, institutional audience, and formal genre. This limits the scope for corpus-level genre differences to explain systematic differences among analyst clusters; remaining associations may instead reflect author-level variation.

CRS mandates balanced, non-advocacy prose, and legal policy analysis generally uses a neutral, non-partisan register. \textit{Sentiment} should therefore be near-zero; a strong correlation would instead suggest analyst-level differences in subtle affective framing despite these institutional constraints.

\textit{Text length} is the most plausible non-trivial confound. Analysts who primarily handle brief overview reports will have shorter documents than those who specialise in comprehensive legislative histories. A strong association between length and UMAP position would indicate that the authorship clusters partly capture specialisation and report scope. Burrows' Delta would nevertheless remain interpretable as authorship evidence because it normalises frequencies rather than raw counts.

\textit{Mean sentence length} and \textit{mean word length} control for syntactic complexity. Although legal writing generally employs complex sentences, analysts may differ in preferred sentence architecture, for example by preferring semicolons to full stops or enumerated lists to long paragraphs. Author-consistent patterns in these measures would support a stylometric interpretation.

In legal writing, \textit{passive-voice density} may reflect the use of passive constructions to hedge or distance the author from normative claims. Analyst-level variation could capture individual style, differential use of legal conventions, or both. An association with the analyst clusters would therefore help specify what the stylometric features capture.

The \textit{coordination/subordination ratio} captures another syntactic preference: coordinate structures (``X and Y and Z'') versus embedded subordinate clauses. Character trigrams can encode this distinction, so moderate correlations aligned with the analyst clusters would support the stylometric interpretation.

\textit{Observed pattern.}
Text length ($|r| \leq 0.11$), mean sentence length ($|r| \leq 0.11$), and mean word length ($r_{\text{U1}} = -0.20$) have weak associations with the projection, so these measures account for little of the analyst-level cluster structure.
The largest correlations involve the coordination/subordination ratio ($r_{\text{U1}} = -0.50$), dictionary sentiment ($r_{\text{U2}} = -0.40$), and passive-voice density ($r_{\text{U2}} = -0.34$).
Because every document shares topic, audience, and institutional writing conventions, these patterns are consistent with analyst-level differences in syntactic architecture (coordinate versus subordinate constructions) and affective framing. Passive-voice density may reflect differential use of the legal convention of hedging as well as individual style.
The sentiment values should be read with some caution, as the \texttt{bing} dictionary's coverage of specialized legal vocabulary is limited.
In this controlled corpus, the weak topic- and length-related associations provide discriminant validity for the authorship interpretation, although they do not establish individual analyst style as the only possible explanation.

\begin{figure}[!htb]
  \centering
  \includegraphics[width=\linewidth]{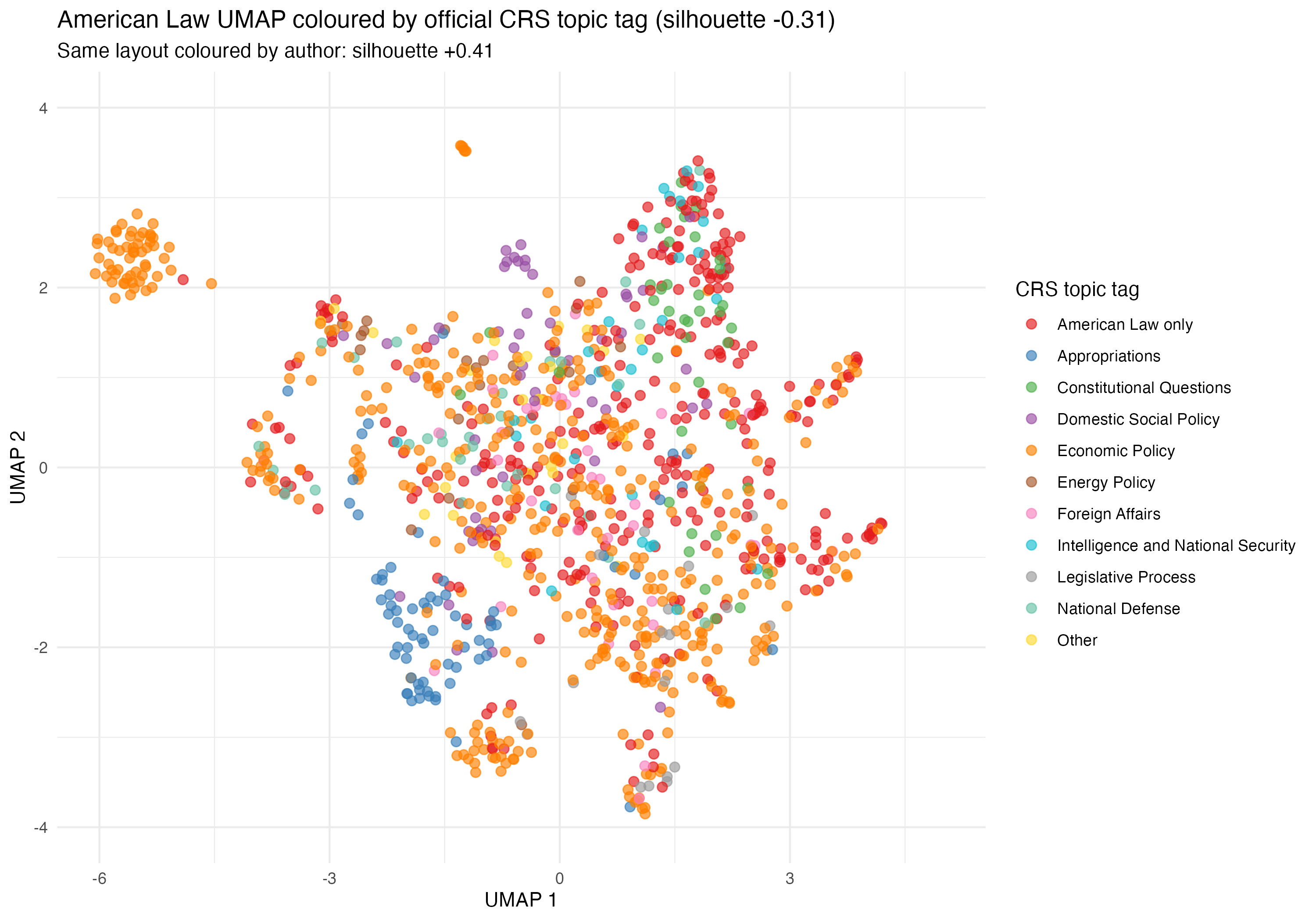}
  \caption{Topic-versus-authorship control for the full CRS American Law subset. The identical character-3-gram UMAP of Figure~\ref{fig:dv_crs}, re-coloured by official CRS topic tags. The topic partition is heavily interspersed (silhouette $-0.31$), against $+0.41$ for author labels in the same coordinates.}
  \label{fig:topic_crs_tags}
\end{figure}

\textit{Topic versus authorship.}
Analyst specialisation within American Law remains a possible confound because analysts have distinct beats. We test this by re-colouring the \emph{identical} character-3-gram UMAP by the official CRS topic tags (Figure~\ref{fig:topic_crs_tags}). Each report's multi-label tag set is collapsed to its first tag other than \emph{American Law}; reports carrying only the division tag form their own category, and tags with fewer than ten reports are grouped as \emph{Other}. The topic partition is heavily interspersed (silhouette $-0.31$), compared with $+0.41$ for author labels in the same coordinates. Subject matter and authorship are moderately associated, as expected when analysts specialise (normalised mutual information $0.40$), but the topic labels do not reproduce the spatial geometry. If topic drove the visible clusters, the topic partition would also form spatially coherent groups. These results are more consistent with authorship than subject matter as the main source of the observed structure.

\textit{Author portability across topics.}
The partition test fixes the projection and varies the labels. A complementary analysis fixes the author and varies the topic. Across the full CRS corpus beyond American Law, 36 analysts each wrote at least six single-authored reports in each of two or more distinct primary topic areas, yielding 942 documents. We compute pairwise cosine similarities between the documents' character-3-gram profiles, using the same tokenization and trimming as the main pipeline, and average them by pair type (Table~\ref{tab:author_topic_portability}). Reports are more similar to reports by the same author on \emph{different} topics ($0.83$) than to reports by other analysts on the \emph{same} topic ($0.72$). The same-author similarity is nearly unchanged across topics ($0.83$, compared with $0.82$ within topic), while the topic effect over the unrelated-pair baseline is $+0.02$, compared with an author effect of $+0.13$. A nearest-neighbour analysis yields a similar pattern. When each report's nearest neighbour is restricted to documents outside its own author--topic cell, the neighbour shares the author in 59 percent of cases and the topic in 16 percent. These results indicate that similarities associated with analyst identity persist across topic areas.

\begin{table}[!htb]
\centering
\small
\begin{threeparttable}
\caption{Author portability across topics: mean pairwise cosine similarity of character-3-gram profiles, CRS analysts writing in multiple topic areas ($n = 942$ reports, 36 analysts, 18 topics)}
\label{tab:author_topic_portability}
\begin{tabular}{@{}lrr@{}}
\toprule
\textbf{Pair type} & \textbf{Mean cosine} & \textbf{Pairs} \\
\midrule
Same author, different topic & $0.83$ & $8{,}526$ \\
Same author, same topic      & $0.82$ & $9{,}193$ \\
Same topic, different author & $0.72$ & $50{,}336$ \\
Different author, different topic & $0.70$ & $375{,}156$ \\
\bottomrule
\end{tabular}
\begin{tablenotes}
\footnotesize
\item Analysts qualify with at least six single-authored reports in each of two or more primary EveryCRSReport topic areas.
\end{tablenotes}
\end{threeparttable}
\end{table}

\subsection{Use Case 3: Hungarian Ombudsman Reports}

\begin{figure}[!htb]
  \centering
  \includegraphics[width=\linewidth]{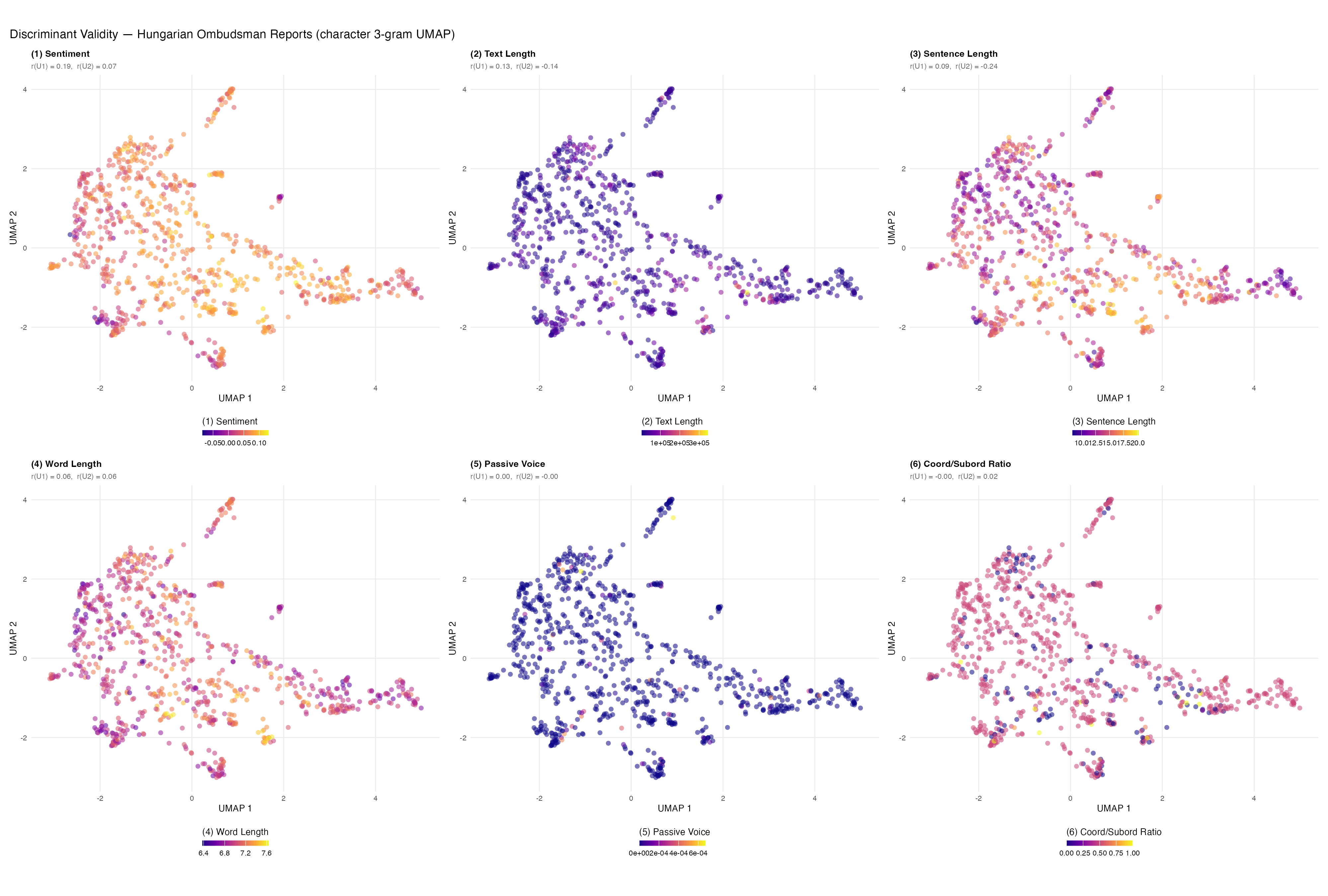}
  \caption{Discriminant validity for Hungarian Ombudsman Reports (Use Case 3).
  Character 3-gram UMAP re-coloured by six alternative factors.
  \textit{Caution}: the syntactic factors use English-language heuristics applied to Hungarian text; quantitative values should be interpreted with care.
  Pearson $r$ with each UMAP dimension is printed as a panel subtitle.}
  \label{fig:dv_ombudsman}
\end{figure}

\begin{figure}[!htb]
  \centering
  \includegraphics[width=\linewidth]{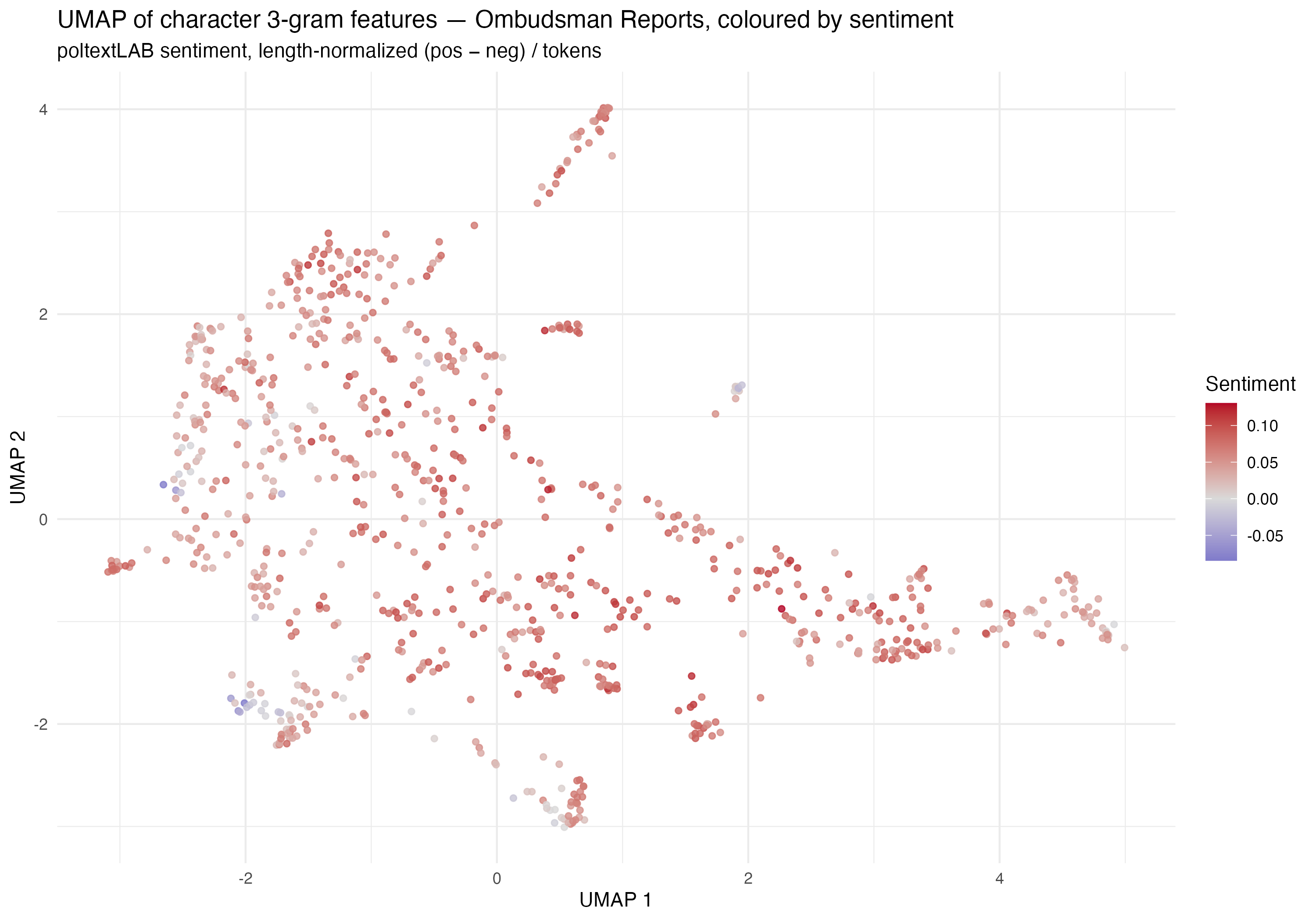}
  \caption{Hungarian Ombudsman reports: the character 3-gram UMAP of the main text, re-coloured by length-normalised \emph{poltextLAB} sentiment. No visible affective gradient organises the layout, making sentiment an unlikely explanation for the rapporteur-level structure.}
  \label{fig:ombudsman_3grams_sentiment}
\end{figure}

\textbf{Interpretation.}
Unlike the Orb\'{a}n speeches, each Ombudsman report carries a named rapporteur label that provides a ground-truth authorship proxy for validation. The discriminant validity analysis assesses whether the rapporteur-level UMAP structure reflects individual style or report-level confounds.

\textit{Sentiment} should be minimally informative because Ombudsman reports are legal-administrative texts written in a highly standardised, neutral register, with little affective vocabulary and limited variation across rapporteurs. Re-colouring the identical projection by dictionary sentiment, computed with the Hungarian-language \emph{poltextLAB} political sentiment lexicon \citep{ring2024}, shows no visible affective gradient organising the layout (Figure~\ref{fig:ombudsman_3grams_sentiment}). The low sentiment--UMAP association supports the authorship interpretation and makes dictionary sentiment an unlikely explanation for the structure.

\textit{Text length} may correlate moderately if different rapporteurs handle complaints of different complexity or scope. All reports share the same formal structure, which limits corpus-level genre variation. A strong association would nevertheless suggest that UMAP is primarily recovering report complexity rather than authorship.

\textit{Mean sentence length} controls for differences in sentence architecture. Legal prose often uses long, subordinate-clause-heavy sentences; concentration of such sentences within one rapporteur's documents would support the authorship interpretation, whereas a pattern crossing rapporteur boundaries would suggest a non-authorship confound.

In Hungarian, \textit{mean word length} reflects morphology as well as style. Case types that require longer compound terms or more suffixes could make word length vary by rapporteur, producing a topic-related rather than author-related signal.

Because these syntactic measures rely on English-language heuristics, \textit{passive-voice density} and the \textit{coordination/subordination ratio} should be treated as rough proxies.

\textit{Observed pattern.}
All six factors correlate weakly with both UMAP dimensions. The largest correlations are for mean sentence length ($|r| \leq 0.24$) and dictionary sentiment ($r_{\text{U1}} = 0.19$, computed here with the Hungarian \emph{poltextLAB} lexicon). The correlations for text length ($|r| \leq 0.14$), mean word length ($|r| \leq 0.06$), passive-voice density ($|r| \leq 0.01$), and the coordination/subordination ratio ($|r| \leq 0.02$) are near zero.
In this corpus of legal-administrative Hungarian reports, none of the six surface-level factors explains much of the rapporteur-level cluster structure visible in the main-text UMAP.
This makes text length, syntactic complexity, and writing register unlikely to drive the observed separation and is consistent with rapporteur-level stylometric differences.

\subsection{Use Case 4: Trump Tweets (Android vs.\ iPhone)}

\begin{figure}[!htb]
  \centering
  \includegraphics[width=\linewidth]{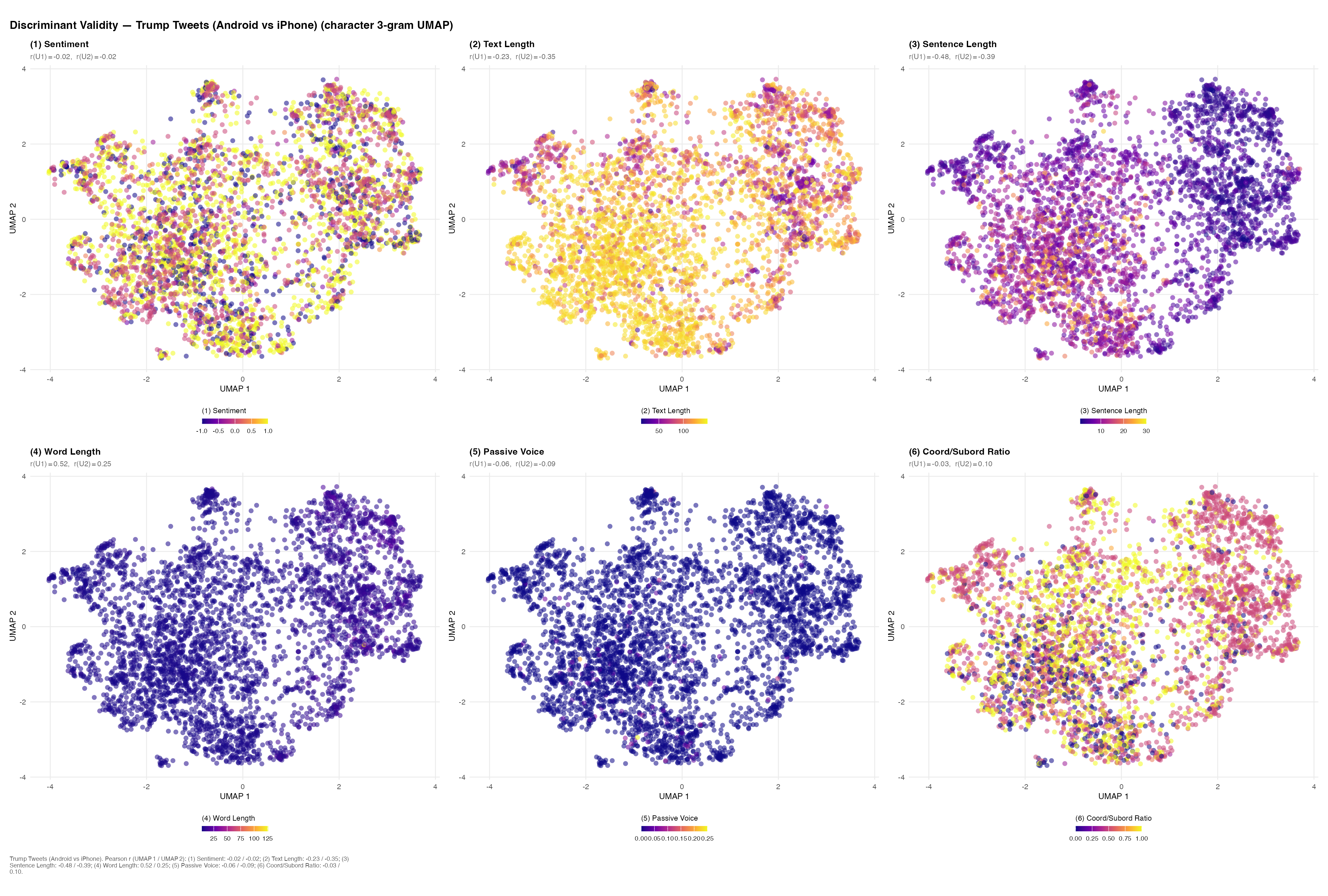}
  \caption{Discriminant validity for Trump Tweets (Use Case 4).
  The same character 3-gram UMAP as in the main text, re-coloured by six alternative factors.
  Pearson $r$ with each UMAP dimension is printed as a panel subtitle.}
  \label{fig:dv_tweets}
\end{figure}

\textbf{Interpretation.}
In this case, surface-form variables are expected to align with the device separation. The Android/iPhone contrast in the main text is organised by production mode: the iPhone stream contains links, hashtags, and promotional posts, whereas the Android stream contains composed prose. Because production register co-varies with length and surface form, correlations with length-related variables would support a register interpretation rather than a clean authorship split.

\textit{Observed pattern.}
Mean word length ($r_{\text{U1}} = 0.52$), mean sentence length ($r_{\text{U1}} = -0.48$), and text length ($r_{\text{U1}} = -0.23$, $r_{\text{U2}} = -0.35$) correlate moderately with the device-separating axes. Dictionary sentiment ($r_{\text{U1}} = -0.02$, $r_{\text{U2}} = -0.02$), passive-voice density ($r_{\text{U1}} = -0.06$, $r_{\text{U2}} = -0.09$), and the coordination/subordination ratio ($r_{\text{U1}} = -0.03$, $r_{\text{U2}} = 0.10$) are effectively zero. This differs from the formal-prose corpora, where length-related factors were null and the recovered structure was authorial. Here those factors track the projection because the device label represents production register. The formatting-stripping analysis provides a direct test: removing URL characters reduces the mean device silhouette from $0.21$ to $0.09$, and removing hashtag characters as well reduces it to $0.08$. This indicates that formatting accounts for most of the device separation.

\textbf{Burrows' Delta, leave-one-out.}
Each device profile is the mean frequent-word vector for that device, so every tweet contributes marginally to its own reference profile (self-weight $1/1{,}866$ for Android, $1/2{,}134$ for iPhone). We recomputed all scores against leave-one-out profiles, comparing each tweet only with profiles built from the \emph{other} tweets. This recomputation changes the mean device scores by under half a percent and leaves the device ranking and separation reported in the main-text table unchanged. The frequent-word contrast is therefore not an artefact of self-inclusion, but it still concerns a label that the stripping ladder identifies with production register rather than authorial hand.

\subsection{Use Case 5: Trump Speeches (Teleprompter vs.\ Off-the-Cuff)}

\begin{figure}[!htb]
  \centering
  \includegraphics[width=\linewidth]{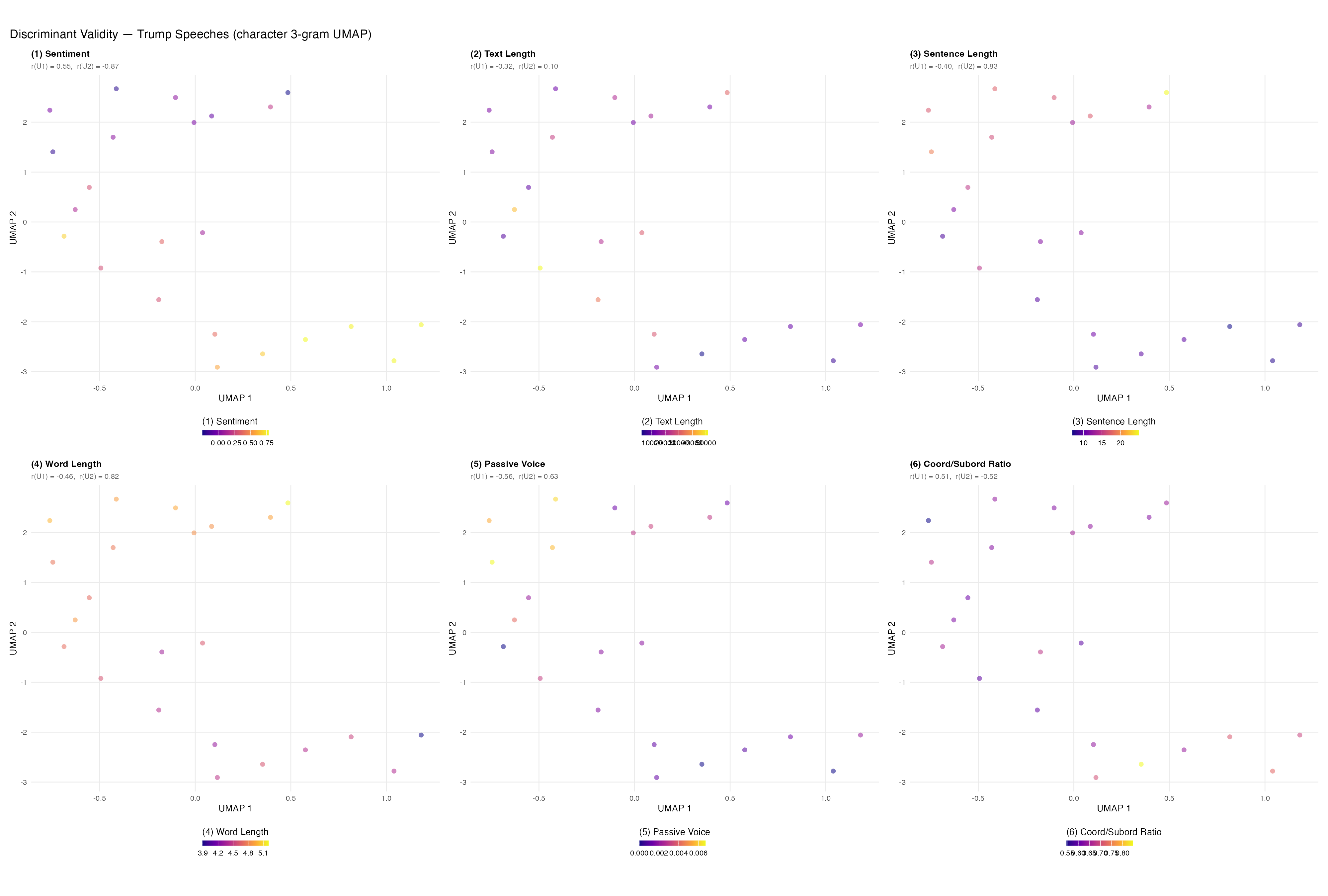}
  \caption{Discriminant validity for Trump Speeches (Use Case 5).
  Character 3-gram UMAP re-coloured by six alternative factors.
  Pearson $r$ with each UMAP dimension is printed as a panel subtitle.}
  \label{fig:dv_speeches}
\end{figure}

\textbf{Interpretation.}
The speech corpus differs from the tweet corpus. Its primary contrast is between two delivery modes by the same speaker: scripted teleprompter delivery and improvised ad-lib delivery, rather than between two individuals.

\textit{Sentiment} differs significantly in the main text, with impromptu speeches scoring higher, so moderate correlations are expected. We assess whether this difference fully explains the UMAP separation. Burrows' Delta on function words, which carry little affective content, also separates the two groups and therefore weighs against a purely sentiment-driven account.

\textit{Text length} is unlikely to drive the separation. The main text reports no statistically significant length difference between teleprompter and off-the-cuff speeches, and duration depends more on event format than delivery style.

\textit{Mean sentence length} captures a plausible fluency and complexity difference. Scripted text tends to use longer, more syntactically complex sentences, while spontaneous speech contains shorter bursts and false starts. A strong correlation would mean that sentence structure accounts for much of the UMAP separation; a moderate one would indicate a contribution without dominance.

\textit{Mean word length} may correlate negatively with the impromptu cluster because unscripted remarks use more colloquial vocabulary, including shorter and more common words. Such a pattern would show that word choice varies with delivery mode.

\textit{Passive-voice density} should be higher in scripted policy language than in ad-lib remarks. A strong alignment with the projection would indicate that formal and informal register account for much of the separation.

The \textit{coordination/subordination ratio} captures a related syntactic difference. Scripted text tends to use subordinating structures for precision (conditional clauses, relative clauses, embedded complements), whereas spontaneous speech favours simpler coordination (``and,'' ``but,'' ``or''). Alignment with the teleprompter/ad-lib split would support a register interpretation; weak correlations would indicate that the delivery-mode difference extends beyond these surface measures.

\textit{Observed pattern.}
The speech corpus has the highest correlations in the study. Sentiment ($r_{\text{U1}} = 0.55$, $r_{\text{U2}} = -0.87$), mean sentence length ($r_{\text{U1}} = -0.40$, $r_{\text{U2}} = 0.83$), and mean word length ($r_{\text{U1}} = -0.46$, $r_{\text{U2}} = 0.82$) align strongly with the UMAP axes.
Passive-voice density ($r_{\text{U1}} = -0.56$, $r_{\text{U2}} = 0.63$) has moderate correlations, and the coordination/subordination correlations are at most $0.52$ in absolute value.
Text length remains low ($r_{\text{U1}} = -0.32$, $r_{\text{U2}} = 0.10$), consistent with the non-significant length difference reported in the main text.
These results are expected because the scripted/impromptu distinction partly concerns register: scripted text has longer sentences, more formal vocabulary, and denser passive constructions.
However, Burrows' Delta on function words, which carry minimal affective or syntactic-complexity signal, produces near-perfect separation between the two groups (see the corresponding figure in the main text). Function-word differences therefore remain alongside the measured register markers.
The discriminant validity figures show that surface register explains part, but not all, of the structure.

\subsection{Use Case 6: Orb\'{a}n Speeches in Hungarian}

\begin{figure}[!htb]
  \centering
  \includegraphics[width=\linewidth]{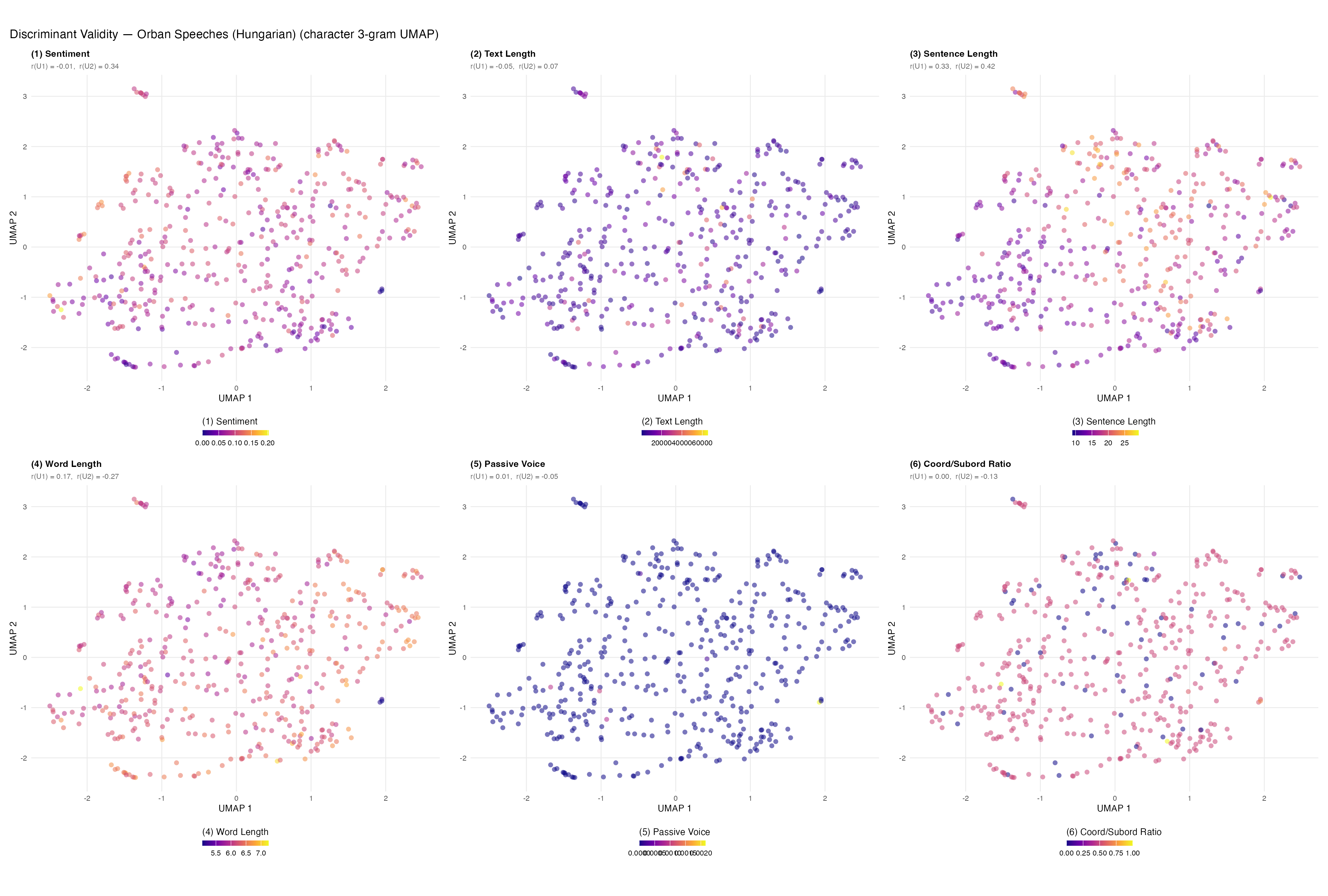}
  \caption{Discriminant-validity battery for 466 delivered speeches by Viktor Orb\'an: the character 3-gram UMAP re-coloured by six non-authorship factors, with the Pearson $r$ against each UMAP dimension printed as a panel subtitle. The layout has no stable clusters across the hyperparameter settings examined. Mean sentence length has the strongest association ($|r_{\text{U2}}| = 0.42$), while sentiment and mean word length correlate more weakly. Sentiment is the length-normalised \emph{poltextLAB} Hungarian score; the syntactic factors use English-language heuristics and are unreliable for Hungarian.}
  \label{fig:dv_orban}
\end{figure}

Because the Orb\'{a}n corpus has no authorship labels, the discriminant-validity battery provides the primary diagnostic (Figure~\ref{fig:dv_orban}). The figure shows the character 3-gram UMAP re-coloured by six non-authorship factors; panel subtitles report Pearson $r$ for each UMAP dimension. This section explains the factors in detail. \textit{Caution}: the syntactic factors are computed using heuristics designed for English; their values for Hungarian text should be interpreted with caution.

\textbf{Interpretation.}
In the Orb\'{a}n corpus, the character 3-gram UMAP in the main text is amorphous and non-clustered. We interpret this as a failure of frequency-based stylometry to recover author-level structure in this setting.

\textit{Sentiment} may distinguish emotionally charged domestic or programmatic speeches from brief international statements. An association with a UMAP axis would indicate that affect explains part of the variation.

\textit{Text length} is a plausible organising factor because the corpus mixes short ceremonial remarks with long ideological keynotes (e.g., the annual \emph{\'{e}v\'{e}rt\'{e}kel\H{o}} addresses and the Tusn\'{a}df\"{u}rd\H{o} lectures). This range could affect a projection that is sensitive to vocabulary richness, including one based on character trigrams.

\textit{Mean sentence length} captures differences in formality: programmatic speeches have longer, more complex sentences, while ceremonial and bilateral statements use shorter, more formulaic ones.

Hungarian's agglutinative morphology makes \textit{mean word length} relevant. Compound words and case suffixes produce longer tokens than their English equivalents, and variation across speech types may reflect topic or genre rather than authorial preference.

\textit{Passive-voice density} and \textit{coordination/subordination ratio} are both computed using English-language heuristics and are unreliable for Hungarian.
Any correlations should be treated as artefacts of tokenization and pattern-matching on a morphologically rich language, not as substantive stylometric signals.

\textit{Observed pattern.}
Mean sentence length ($r_{\text{U2}} = -0.42$) has the strongest correlation, followed by dictionary sentiment ($r_{\text{U2}} = -0.30$, length-normalised \emph{poltextLAB}) and mean word length ($r_{\text{U1}} = -0.27$).
These moderate correlations suggest that speech formality and genre (keynote vs.\ ceremonial address), rather than authorship, weakly organise the projection.
Text length ($r_{\text{U1}} = 0.05$, $r_{\text{U2}} = -0.05$) is near zero despite the wide length range of the corpus, suggesting that UMAP is not simply sorting speeches by duration.
Passive voice and the coordination/subordination ratio are also near zero, as expected because the English-language heuristics used to compute them have limited validity for Hungarian morphology.
No single factor dominates the projection, but register and sentence-level formality weakly shape it. This is consistent with the main text's conclusion that frequency-based stylometry fails to recover stable authorship structure in this setting.

\bibliography{references_v2}

\end{document}